\documentclass[10pt,journal]{IEEEtran}
\ifCLASSINFOpdf
\else
\fi
\usepackage{graphicx}
\ifCLASSOPTIONcompsoc
  \usepackage[nocompress]{cite}
\else
  \usepackage{cite}
\fi
\usepackage{amssymb}
\usepackage{amsmath}
\usepackage{multirow}
\usepackage{amsmath}
\usepackage{amssymb}
\usepackage{booktabs}
\usepackage{algorithm}
\usepackage{algorithmic}
\usepackage{color}
\usepackage[colorlinks,linkcolor=black, urlcolor=black,citecolor=black]{hyperref}

\begin{document}

\title{Revisiting Initializing Then Refining: An Incomplete and Missing Graph Imputation Network}

\author{Wenxuan Tu,
        Bin Xiao,
        Xinwang Liu,~\IEEEmembership{Senior Member, IEEE},
        Sihang Zhou,
        Zhiping Cai,
        and Jieren Cheng

\thanks{W. Tu, X. Liu, and Z. Cai, are with the College of Computer, National University of Defense Technology, Changsha 410073, China (e-mail: \{wenxuantu\}@163.com, \{xinwangliu,zpcai\}@nudt.edu.cn).} 
\thanks{B. Xiao is with the Department of Computer Science and Technology, Chongqing University of Posts and Telecommunications, Chongqing 400065, China  (e-mail: xiaobin@cqupt.edu.cn).}
\thanks{S. Zhou is with the College of Intelligence Science and Technology, National University of Defense Technology, Changsha 410073, China (e-mail: sihangjoe@gmail.com).}
\thanks{J. Cheng is with the School of Computer Science and Technology, Hainan University, Haikou, 570228 China, with the Hainan blockchain technology engineering research center, Haikou 570228, China (E-mail: cjr22@163.com).}

}

\markboth{ }%
{Shell \MakeLowercase{\textit{Tu et al.}}: Revisiting Initializing Then Refining: An Incomplete and Missing Graph Imputation Network}
\maketitle

\begin{abstract}
With the development of various applications, such as social networks and knowledge graphs, graph data has been ubiquitous in the real world. Unfortunately, graphs usually suffer from being absent due to privacy-protecting policies or copyright restrictions during data collection. The absence of graph data can be roughly categorized into attribute-incomplete and attribute-missing circumstances. Specifically, attribute-incomplete indicates that a part of the attribute vectors of all nodes are incomplete, while attribute-missing indicates that the whole attribute vectors of partial nodes are missing. Although many graph imputation methods have been proposed, none of them is custom-designed for a common situation where both types of graph data absence exist simultaneously. To fill this gap, we develop a novel graph imputation network termed Revisiting Initializing Then Refining (RITR), where we complete both attribute-incomplete and attribute-missing samples under the guidance of a novel initializing-then-refining imputation criterion. Specifically, to complete attribute-incomplete samples, we first initialize the incomplete attributes using Gaussian noise before network learning, and then introduce a structure-attribute consistency constraint to refine incomplete values by approximating a structure-attribute correlation matrix to a high-order structural matrix. To complete attribute-missing samples, we first adopt structure embeddings of attribute-missing samples as the embedding initialization, and then refine these initial values by adaptively aggregating the reliable information of attribute-incomplete samples according to a dynamic affinity structure. To the best of our knowledge, this newly designed method is the first end-to-end unsupervised framework dedicated to handling hybrid-absent graphs. Extensive experiments on four datasets have verified that our methods consistently outperform existing state-of-the-art competitors. 
\end{abstract}


\maketitle

\section{Introduction}
\IEEEPARstart{G}{raphs}, which model and represent the complicated relationships among real-world objects, are ubiquitous in practical scenarios, including citation graphs, social graphs, protein graphs, and molecule graphs. To analyze the graph data, graph machine learning attempts to transform an original graph into low-dimensional sample representations by preserving node attributes and graph structure simultaneously \cite{2022NDP, 2022AdvCaching, 2022MT-MVGCN}. In recent years, with the help of graph neural networks (GNNs), graph machine learning has become an increasingly powerful artificial intelligence technique. It has achieved significant success in diverse real-world applications, such as anomalous citation detection \cite{2022GLAD}, few-shot learning \cite{2022HGNN}, feature selection \cite{2022EGCFS}, and knowledge graphs \cite{2022HRAN}.

\begin{figure}[!t]
		\centering
		\includegraphics[width=3.5in]{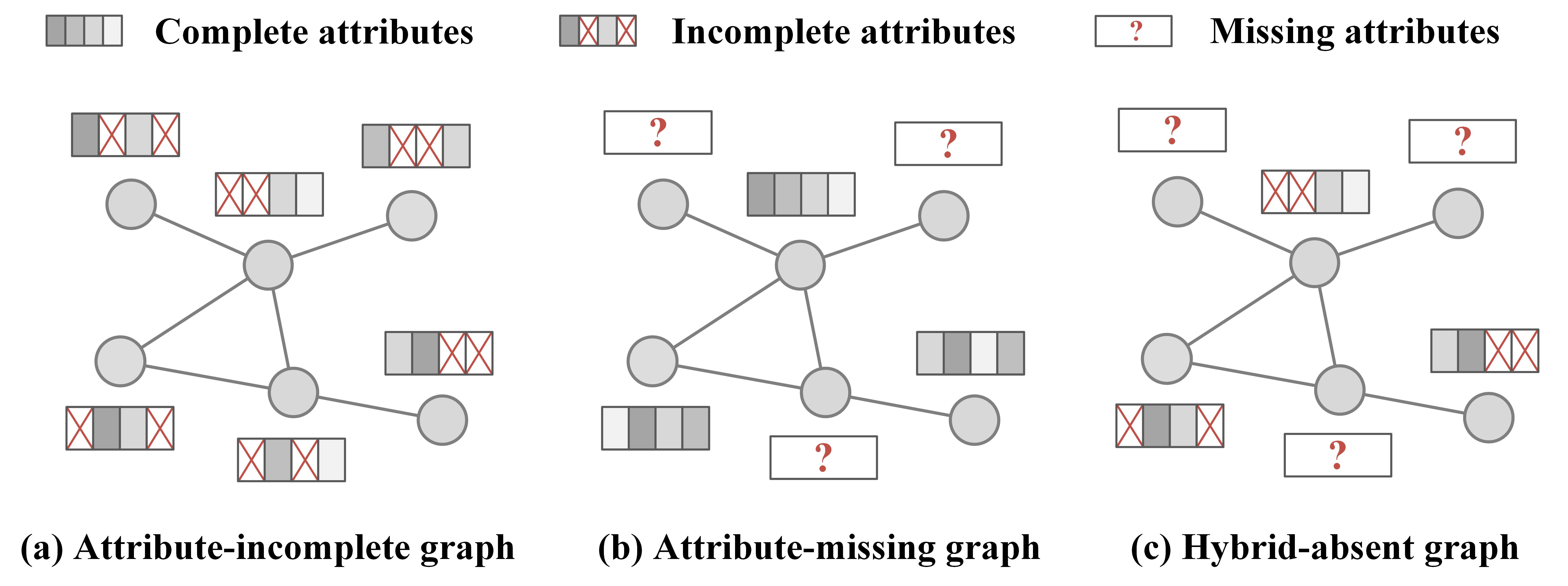}
		\caption{Different types of absent graphs. (a) Attribute-incomplete graph: 
		particular attributes of all samples are absent; (b) Attribute-missing graph: all attributes of specific nodes are absent; (c) Hybrid-absent graph: both circumstances (a) and (b) exist simultaneously within a graph. As one can easily see, the last category is the most challenging, however, it is still under-explored in previous literature. We make the first attempt to solve it by proposing a novel method called RITR.}
		\label{1}
\end{figure}

The key prerequisite for the impressive performance of existing graph machine learning methods lies in the assumption that all samples within a graph are available and complete. However, this assumption may not always hold in practice since it is hard to collect all information from graph data. The reason behind including but not limited to privacy-protecting policies, copyright restrictions, and simply not enough information. For example, in a co-purchase graph of Amazon, consumers tend to selectively (or entirely not) provide their feedback for specific items due to privacy concerns. In a citation network, some papers are inaccessible due to copyright protection. All these circumstances could easily trigger sparsity and data-absent problems that adversely affect graph representations. According to the type of node attribute absence, the absent graphs can be roughly divided into two categories: 1) the attribute-incomplete graph where only a portion of attributes of all nodes are absent; 2) the attribute-missing graph where all attributes of specific nodes are absent. Fig. \ref{1} illustrates the situations of attribute absence. Among them, Fig. \ref{1}(a) corresponds to an attribute-incomplete graph, Fig. \ref{1}(b) corresponds to an attribute-missing graph, and Fig. \ref{1}(c) corresponds to a hybrid-absent graph where both attribute-incomplete samples and attribute-missing samples exist in the same graph. The above cases make valuable information invisible and pose significant challenges to existing graph machine learning methods for graph analysis.

To solve the attribute-incomplete learning problem, many efforts have been devoted to developing various imputation strategies such as matrix completion \cite{2017GCMC,2020IGMC}, generative adversarial network \cite{2020GINN}, Gaussian mixture model (GMM) \cite{2021GCNMF}, and other advanced ones \cite{2021Inductive,2022From}. With the imputed attributes, these methods then integrate a standard GNN-based framework with data imputation techniques to conduct the sample embedding. Although significant progress has been made in solving the attribute-incomplete learning problem, the performance of these methods degrades drastically when they handle extremely absent data (\textit{e.g.,} attribute-missing graphs). To solve this problem, a recent advanced method termed SAT \cite{2020SAT} first introduces an unsupervised graph imputation framework to handle attribute-missing graphs under the guidance of a shared-latent space assumption. 
Specifically, SAT utilizes a graph neural network (\textit{e.g.,} GCN \cite{2017Semi} or GAT \cite{2018GAN}) to embed the available attributes and graph structure into a latent space in a decoupled manner. Then it performs a distribution matching mechanism to recover the unknown values of attribute-missing samples. Although achieving encouraging success, SAT also suffers from the following limitations when conducting the data imputation: 1) two-source information isolation. SAT isolates the learning processes of embeddings of observed attributes and the complete graph structure. This prevents the trustworthy visible information from being sufficiently utilized, which could cause the learned representations to be biased and also increase the risk of inaccurate data imputation; 2) strict prior assumption. SAT forces two-source latent variables to align with an in-discriminative noise matrix obeying normal distribution. While in reality, the pre-defined normal distribution would not ideally conform to the complex graphs. As a result, the negotiation between attribute and structure information tends to get overly rigid, resulting in less discriminative representations. This could adversely affect the quality of the rebuilt attribute matrix of all samples, especially those without attributes.
 
To overcome the above issues, we propose an $\textbf{\underline{I}}$nitializing $\textbf{\underline{T}}$hen $\textbf{\underline{R}}$efining (ITR) network \cite{2022ITR} to forbid the adverse effect of inaccurate simple initialization and the limitation of rigid distribution assumption. The core idea of ITR is to fully leverage the trustworthy visible information to implement the sample embeddings for missing attribute imputation. Though being the potential for better tackling the attribute-missing problem, we observe that when ITR processes hybrid-absent graphs, attribute-incomplete samples would largely undermine the quality of generated attribute-missing features due to the diffusion of inaccurate imputed information. To our knowledge, hybrid-absent graph machine learning has not been studied in the existing graph literature, which is a universal and more challenging problem for various practical applications. To fill this gap, we revisit ITR and further improve it by designing a variant, termed $\textbf{\underline{R}}$evisiting $\textbf{\underline{I}}$nitializing $\textbf{\underline{T}}$hen $\textbf{\underline{R}}$efining (RITR). In this newly proposed RITR, we complete both attribute-incomplete and attribute-missing samples under the guidance of the initializing-then-refining imputation criterion. 

To impute incomplete attributes, we elaborately design a $\textbf{\underline{S}}$ample-denoising $\textbf{\underline{T}}$hen $\textbf{\underline{C}}$onsistency-preserving (STC) mechanism. As illustrated in Fig. \ref{2}, the feature completion process within this component mainly includes step \textit{1} to step \textit{3}. Firstly, we learn the sample embeddings through a denoising learning approach by combining the attribute and structure information of attribute-incomplete samples sufficiently. Secondly, we take all nodes and learn sample embeddings only according to the structure information. Finally, we explicitly develop a structure-attribute consistency constraint to refine these incomplete latent variables by approximating a structure-attribute correlation matrix to a high-order structure matrix. This operation aims to guarantee the representation quality of nodes with incomplete attributes, and meanwhile provide a feasible initialization to those nodes with missing attributes in the next step. 
To impute missing attributes, we design another data imputation mechanism termed $\textbf{\underline{I}}$nitializing $\textbf{\underline{T}}$hen $\textbf{\underline{R}}$efining (ITR). In this component, we first take the structure embeddings of attribute-missing samples as initial imputed variables and then refine them with an adaptively updated affinity structure for embedding refinement. The above operations correspond to step \textit{4} and step \textit{5} in Fig. \ref{2}. 
Comprehensive experiments on four benchmark datasets have been conducted to verify the effectiveness and superiority of our proposed methods and components. As demonstrated, ITR consistently outperforms state-of-the-art methods. Moreover, the other proposed variant, \textit{i.e.,} RITR, further improves the profiling and classification performance against ITR. It is expected that the simplicity and effectiveness of the RITR method will make it a promising option to be considered for practical applications where the hybrid-absent case is encountered.


This work is a substantially extended version of our original conference paper \cite{2022ITR}. Compared with its previous version, it has the following significant improvements. 1) \textit{\textbf{Novel research problem}}. To the best of our knowledge, hybrid-absent graph machine learning is a rarely explored research field yet is a real-world demand from various applications. Accordingly, we develop a novel graph machine learning framework called RITR without relying on any pre-assumed distribution assumption, which is the first incomplete and missing graph imputation network to solve the corresponding learning problem.
2) \textit{\textbf{Newly proposed strategy}}. To complete attribute-incomplete samples, we propose a new feature completion mechanism termed STC by following the initializing-then-refining imputation criterion. This operation enables the model to generate more robust and discriminative features for attribute-incomplete samples, best serving subsequent attribute-missing imputation tasks.
3) \textit{\textbf{More experimental results and analyses}}. Besides more detailed discussion and extension, we also conduct more comprehensive experiments, and all evaluation results have verified that the two proposed methods achieve the best performance in different absent graph situations.

The remainder of this paper is organized as follows. Section 2 reviews related work in terms of unsupervised graph machine learning and graph machine learning on absent graphs. Section 3 presents the notations, definitions, network and component design, and learning targets. Section 4 conducts experiments and discusses the results. Section 5 draws a final conclusion.

\section{Related Work}
\subsection{Unsupervised Graph Machine Learning} 
Early solutions to unsupervised graph machine learning mainly focus on random walk-based methods \cite{2014DeepWalk,2016node2vec,2018UniWalk}, which first generate the random walk sequences over the network structure properties and then utilize a Skip-Gram model to learn graph representations. However, these methods heavily rely on structure information and overlook other available properties (\textit{e.g.,} attribute information) in the graph. More recently, since the powerful neighborhood aggregation capacity of graph neural networks (GNNs), many efforts have been made to design GNN-based methods. As one of the most representatives, generative/predictive learning-oriented methods aim to explore abundant information embedded in data via some well-known approaches, such as auto-encoder learning \cite{2016Variational, Bo2020Structural,2021DFCN, DCRN} and adversarial learning \cite{2019Adversarial,2021GraphAdversarial,2021Learning,2020ARGA}. Another line pays attention to graph contrastive learning, which aims to maximize the agreement of two jointly sampled positive pairs \cite{2019Deep, 2020Contrastive, 2020Graph, 2021GCA, 2022AGC,2022MVDGI}. One underlying assumption commonly adopted by these methods is that the attributes of all nodes are trustworthy and complete. While in real-world scenarios, they may suffer from significant performance degradation when handling absent graphs.

\subsection{Graph Machine Learning on Absent Graphs}
According to the type of absent graphs, we can roughly group existing absent graph machine learning methods into the following three categories.  
\subsubsection{Attribute-incomplete Graph Machine Learning} 
In the attribute-incomplete circumstance, some methods propose to leverage data imputation-oriented techniques to restore the incomplete information, such as matrix completion \cite{2017GCMC}, generative adversarial network (GAN) \cite{2014GAN}, and Gaussian mixture model (GMM) \cite{2006Pattern}. For instance, NMTR \cite{2019NMR} and GRAPE \cite{2020GRAPE}, two typical matrix completion methods, first take the user-item rating matrix, users (or items), and the observed ratings as a bipartite graph, sample attributes, and connected relationships, respectively. These methods then adopt a graph neural network to predict the probabilities (regarded as imputed values) of absent connected relationships. Similarly, GRAPE \cite{2020GRAPE} first converts a data imputation task into a linkage prediction learning process over a created bipartite graph, and then utilizes a graph neural network to solve it. Recent efforts like NMTR \cite{2019NMR} and IGMC \cite{2020IGMC} follow the same paradigm as previous matrix completion methods to conduct the data imputation and sample embedding in a transductive or inductive learning manner. In addition, GINN \cite{2020GINN} first initializes the incomplete values by a binary mask matrix before network training, and then learns a graph neural network with an adversarial learning mechanism to complete the absent information. GCNMF \cite{2021GCNMF} utilizes a Gaussian mixture model to estimate the incomplete features according to the available information, and in the meanwhile, jointly optimizes the Gaussian mixture model and graph neural network in a united framework. More recently, T2-GNN \cite{2022T2-GNN} designs a general teacher-student graph learning framework to restore both incomplete node features and graph structure through distillation.

\subsubsection{Attribute-missing Graph Machine Learning}
Compared to the attribute-incomplete circumstance, handling the graph data with a majority of samples having no attributes poses more challenges in learning high-quality node representations. This topic has attracted great attention from graph machine learning researchers recently. For example, HGNN-AC \cite{2021Heterogeneous} first adopts current heterogeneous information networks (HINs) to learn node topological representations, and then utilizes the topological relationship between nodes as guidance to implement feature completion for attribute-missing samples via an attention mechanism. HGCA \cite{2022Analyzing} is an unsupervised heterogeneous graph contrastive learning approach for heterogeneous graphs with missing attributes. It employs the contrastive learning technique to unify the processes of feature completion and representation learning, and thereafter, conduct a fine-grained attribute completion by extracting the semantic relations among different types of samples. Besides the attribute-missing heterogeneous graph machine learning, an advanced method called SAT \cite{2020SAT} makes the first attempt to solve the attribute-missing learning problem over the homogeneous graphs. By unifying the data imputation and network learning processes into a single optimization procedure, SAT learns two-source information embedding matrices in a decoupled manner and then aligns them with a noise matrix sampled from a normal distribution for attribute restoration. Another recent work, ITR \cite{2022ITR} introduces an initializing-then-refining mechanism, enabling the network to fully use the trustworthy visible information to adaptively conduct the sample embedding for missing attribute imputation. More recently, SVGA \cite{2022SVGA} and Amer \cite{2022AMER} develop an auto-encoder-style framework to estimate missing node features via structured variational inference and adversarial learning techniques, respectively.  

\subsubsection{Hybrid-absent Graph Machine Learning}
As aforementioned, attribute-incomplete and attribute-missing graph machine learning problems have been intensively studied in recent years. Despite their significant progress, in nature, these methods are not capable of effectively handling hybrid-absent graphs. In this circumstance, especially for the unsupervised scenario, the performance of existing attribute-incomplete and attribute-missing methods could drop drastically since they suffer from at least one of the following limitations: 1) heavily relying on annotated graph data; 2) lacking a specialized feature completion mechanism for handling attribute-missing (or attribute-incomplete) samples; 3) disconnecting the processes of data imputation and network optimization; 4) isolating the learning processes of structure and attribute embeddings; 5) imposing too strict a distribution assumption on the latent variables. Although our recently proposed ITR could address most of the above problems and exhibits powerful learning capacity in the attribute-missing situation, it is still a great challenge to recover incomplete and missing values with limited available information simultaneously. To achieve this goal, we study a new important research problem termed hybrid-absent graph machine learning, and further improve ITR by designing another variant called RITR. It proposes to first leverage the intimate structure-attribute relationship to guide the imputation of incomplete attributes and then employ the most trustworthy visible information to implement the missing attribute completion. To the best of our knowledge, none of the above literature considers the hybrid-absent graph machine learning problem. RITR is the first work dedicated to this field.

\section{Approach}
\begin{figure*}[!t]
\centering
\includegraphics[width=7.1in]{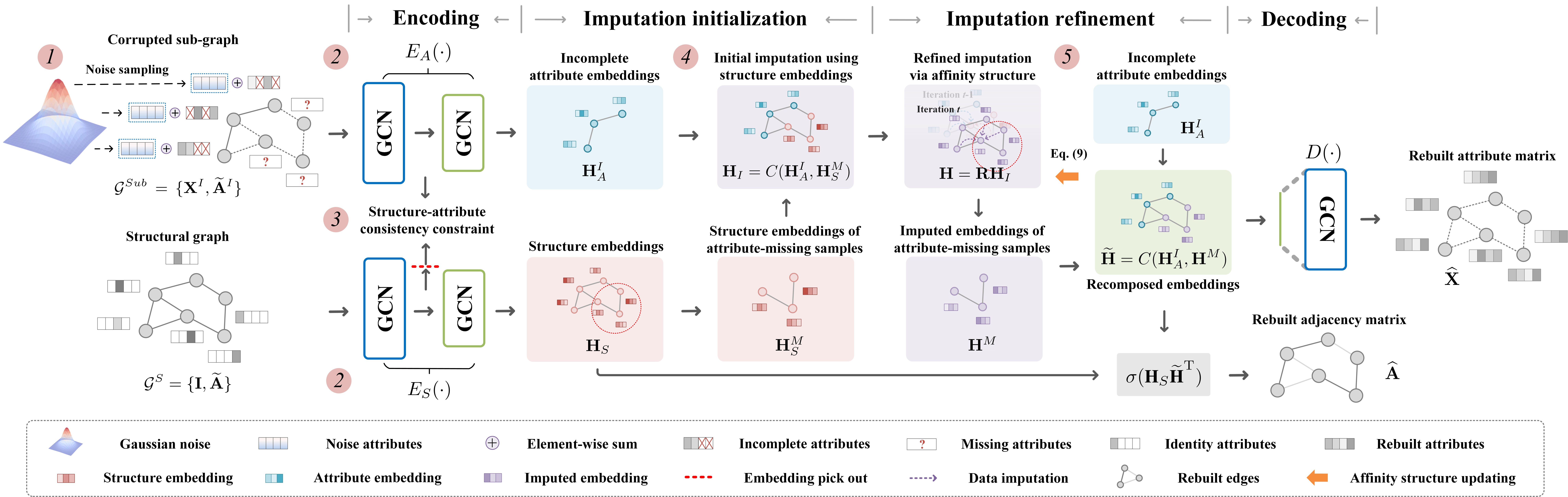}
\caption{The architecture of the revisiting initializing then refining (RITR) framework. To impute the incomplete values, we first initialize the original incomplete attributes as Gaussian noise for denoising learning (\textit{i.e.,} step \textit{1}), and then introduce a structure-attribute consistency constraint to refine the incomplete values by approximating a structure-attribute correlation matrix to a high-order structural matrix (\textit{i.e.,} step \textit{3}). To impute the missing values, we first adopt the structure embeddings of the attribute-missing samples as the embedding initialization (\textit{i.e.,} step \textit{4}), and then adaptively refine these initial values by aggregating the reliable and informative information of the attribute-incomplete samples according to the affinity structure (\textit{i.e.,} step \textit{5}).}
\label{2}
\end{figure*}

Fig. \ref{2} shows an overview of our proposed RITR. As follows, we will provide details on notations, definitions, crucial components, and learning targets, respectively.

\subsection{Notations and Definitions}
$\mathcal{G}=\{\mathcal{V, E}\}$ denotes a given undirected graph that contains $N$ samples with $C$ categories, where $\mathcal{V}$ and $\mathcal{E}$ indicate the node set and edge set, respectively. Generally, the topology of a graph $\mathcal{G}$ can be characterized by its adjacency matrix $\mathbf{A} \in \mathbb{R}^{N \times N}$ and the content of graph $\mathcal{G}$ can be represented by its of attribute matrix $\mathbf{X} \in \mathbb{R}^{N \times D}$, where $D$ refers to the sample dimension. The main notations and their explanations are summarized in Table \ref{I}.

\textbf{Definition 1 (Hybrid-absent Graph)}.
We denote a hybrid-absent graph $\mathcal{\widetilde{G}}=\{\mathcal{V}^I, \mathcal{V}^M, \mathcal{E}\}$, where partial attributes of some samples are unavailable (\textit{i.e.,} the attribute-incomplete sample set $\mathcal{V}^{I}$) and all attributes of other samples are entirely missing (\textit{i.e.,} the attribute-missing sample set $\mathcal{V}^{M}$).
$N^{I}=|\mathcal{V}^{I}|$ and $N^{I}=|\mathcal{V}^{M}|$ refer to the number of attribute-incomplete samples and attribute-missing samples, respectively.
Accordingly, $\mathcal{V}$ = $\mathcal{V}^{I}$ $\cup$ $\mathcal{V}^{M}$, $\mathcal{V}^{I}$ $\cap$ $\mathcal{V}^{M}$ = $\varnothing$, and $N$ = $N^{I}$ + $N^{M}$. Note that the structure information (\textit{i.e.,} $\mathcal{E}$) of $\mathcal{\widetilde{G}}$ is complete. 

\textbf{Definition 2 (Learning Task)}.
In this work, we mainly focus on addressing the hybrid-absent graph machine learning problem on graphs without label annotation. Our auto-encoder-style framework works for learning two graph encoding functions $E_A(\cdot)$ and $E_s(\cdot)$ to impute invisible latent variables. Then a graph encoding function $D(\cdot)$ will recover the attribute-missing and attribute-incomplete samples based on the imputed hidden features. The recovered attributes can be saved and used for profiling and node classification tasks.

\subsection{Overview}
To tackle the issue that existing studies fail to perform well on hybrid-absent graphs, we introduce an end-to-end unsupervised graph imputation network termed RITR. Our goal is to design personalized incomplete and missing feature completion mechanisms on hybrid-absent graphs, and achieve accurate data imputation and effective information propagation with the initializing-then-refining imputation criterion. As illustrated in Fig. \ref{2}, RITR contains two core components, \textit{i.e.,} $\textbf{\underline{S}}$ample-denoising $\textbf{\underline{T}}$hen $\textbf{\underline{C}}$onsistency-preserving (STC) mechanism (Section III-C) and $\textbf{\underline{I}}$nitializing $\textbf{\underline{T}}$hen $\textbf{\underline{R}}$efining (ITR) mechanism (Section III-D), which are intended to solve the attribute-incomplete and attribute-missing problems, respectively. Specifically, before network learning, we first generate a corrupted sub-graph by randomly adding Gaussian noise information to incomplete attributes as initial values. Then the corrupted sub-graph and the structural graph are transferred into two architecture-identical yet decoupled graph encoders to learn low-dimensional representations. In the information extraction phase, a structure-attribute consistency constraint allows the intermediate structure-attribute representations of attribute-incomplete samples to negotiate with each other to refine incomplete attributes.
After that, the ITR mechanism utilizes the structure embeddings of the attribute-missing samples as the embedding initialization, and then adaptively refines these initial values by aggregating the reliable information of attribute-incomplete samples according to an affinity structure. Finally, RITR conducts the structure and attribute reconstructions over the latent embeddings by jointly minimizing three objectives.

\begin{table}[!t]
\centering
\caption{Summary of main notations}
\footnotesize
\begin{tabular}{l|l}
\hline
\multicolumn{1}{c|}{Notation} & \multicolumn{1}{c}{Explanation}   \\\hline
$\mathbf{X} \in \mathbb{R}^{N \times D}$    & Original attribute matrix   \\
$\mathbf{X}^I \in \mathbb{R}^{N^I \times D}$    & Incomplete attribute matrix   \\
$\mathbf{N} \in \mathbb{R}^{N^I \times D}$    & Gaussian noise matrix   \\
$\widetilde{\mathbf{X}}^I \in \mathbb{R}^{N^I \times D}$    & Corrupted incomplete attribute matrix   \\
$\mathbf{A} \in \mathbb{R}^{N \times N}$    & Original adjacency matrix   \\
$\widetilde{\mathbf{A}} \in \mathbb{R}^{N \times N}$    & Normalized adjacency matrix   \\
$\widetilde{\mathbf{A}}^I \in \mathbb{R}^{N^I \times N^I}$    & Normalized adjacency matrix of $\mathcal{G}^{Sub}$    \\
$\mathbf{I} \in \mathbb{R}^{N \times N}$    & Identical matrix  \\
$\mathbf{C} \in \mathbb{R}^{N^I \times N^I}$   & Structure-attribute correlation matrix  \\
$\widetilde{\mathbf{A}}^{Ir} \in \mathbb{R}^{N^I \times N^I}$    & $r$-order normalized adjacency matrix of $\mathcal{G}^{Sub}$    \\
$\mathbf{H}^{I}_{A} \in \mathbb{R}^{N^I \times d}$   & Attribute embeddings of incomplete samples \\
$\mathbf{H}_{S} \in \mathbb{R}^{N \times d}$   & Structure embeddings  \\
$\mathbf{H}^{M}_{S} \in \mathbb{R}^{N^M \times d}$   & Structure embeddings of missing samples  \\
$\mathbf{H}_{I} \in \mathbb{R}^{N \times d}$   &  Initially imputed embeddings  \\
$\mathbf{R} \in \mathbb{R}^{N \times N}$   &  Affinity structure  \\
$\mathbf{H} \in \mathbb{R}^{N \times d}$   &  Imputed embeddings  \\
$\mathbf{H}^M \in \mathbb{R}^{N^M \times d}$   &  Imputed embeddings of missing samples \\
$\widetilde{\mathbf{H}} \in \mathbb{R}^{N \times d}$   &  Sample-recomposed embeddings  \\
$\mathbf{S} \in \mathbb{R}^{N \times N}$   & Normalized self-correlated matrix  \\
$\mathbf{M} \in \mathbb{R}^{N^I \times D}$   & Element-wise indicator matrix   \\
$\widehat{\mathbf{X}} \in \mathbb{R}^{N \times D}$ & Rebuilt attribute matrix   \\
$\widehat{\mathbf{X}}^I \in \mathbb{R}^{N^I \times D}$ & Rebuilt attribute matrix of incomplete samples   \\
$\widehat{\mathbf{A}} \in \mathbb{R}^{N \times N}$ & Rebuilt adjacency matrix   \\
\hline
\end{tabular}

\label{I}
\end{table}

\subsection{Sample-denoising Then Consistency-preserving}
This work attempts to solve an under-explored yet more challenging hybrid-absent graph machine learning task, where attribute-incomplete and attribute-missing samples exist simultaneously within a graph. The critical technique extension of RITR against its conference version is the STC component, which aims to make the learned representations robust to the attribute-incomplete input pattern and discriminative to serve the subsequent attribute-missing imputation task. 

Since the network could not be optimized over unknown values, we employ a sample-denoising learning approach to ease the network training and facilitate the robustness of the learned attribute-incomplete features. Specifically, we first randomly generate a Gaussian noise matrix $\mathbf{N} \in \mathbb{R}^{N^{I} \times D}$ with iterations, and then assign it to the incomplete attribute matrix $\mathbf{X}^I \in \mathbb{R}^{N^{I} \times D}$ as initial values. The resultant matrix is denoted as a corrupted incomplete attribute matrix $\widetilde{\mathbf{X}}^I \in \mathbb{R}^{N^{I} \times D}$. In the attribute-incomplete circumstance, we merely minimize the reconstruction errors of visible attributes between the original attribute matrix and the rebuilt attribute matrix for sample denoising. Although the sample-denoising scheme has been previously explored and proved to be powerful \cite{2008Extracting},
directly applying it to the restoration of incomplete attributes is less comprehensive since partial attributes are invisible. To guarantee the high quality of representations of attribute-incomplete samples, it is intuitive that exploring the complete structure information could benefit the reconstruction of incomplete attributes since both types of latent variables share consistent and complementary properties of a typical graph \cite{2018DANE}. Motivated by this, instead of directly addressing the incompleteness issue in the input space, we leverage the intimate relationship between structure-attribute embeddings to further refine the initial imputation of incomplete attributes.
To be specific, we first utilize two graph convolution network-based encoders denoted as $E_A(\cdot)$ and $E_S(\cdot)$ to extract the latent features of attribute-incomplete samples and graph structure, respectively. Formally, given a sub-graph $\widetilde{\mathcal{G}}^{Sub}$ with a corrupted incomplete attribute matrix $\widetilde{\mathbf{X}}^I$ and a corresponding normalized adjacency matrix $\widetilde{\mathbf{A}}^I \in \mathbb{R}^{N^{I} \times N^{I}}$, $E_A(\cdot)$ accepts them as input and the \textit{l}-th latent representations of attribute-incomplete samples can be formulated as below:
\begin{equation} \label{eq:1}
\mathbf{H}^{I(l)}_{A} = \sigma(\widetilde{\mathbf{A}}^{I}\mathbf{H}^{I(l-1)}_{A}\mathbf{\Theta}^{(l)}),
\end{equation}
where $\mathbf{\Theta}^{(l)}$ is the parameter matrix of $E_A(\cdot)$ in the \textit{l}-th layer and $\sigma(\cdot)$ indicates a non-linear activation function. Similarly, $E_S(\cdot)$ receives a structure graph $\mathcal{G}^{S}$ with an identity matrix $\mathbf{I} \in \mathbb{R}^{N \times N}$ and a normalized adjacency matrix $\widetilde{\mathbf{A}} \in \mathbb{R}^{N \times N}$, and the \textit{l}-th latent representations of the graph structure can be obtained:
\begin{equation} \label{eq:2}
\mathbf{H}^{(l)}_{S} = \sigma(\widetilde{\mathbf{A}}\mathbf{H}^{(l-1)}_{S}\mathbf{\Psi}^{(l)}),
\end{equation}
where $\mathbf{\Psi}^{(l)}$ is the parameter matrix of $E_S(\cdot)$ in the \textit{l}-th layer.
After that, we refine the intermediate information of attribute-incomplete samples via a structure-attribute consistency constraint, so as to make the learned representations much better suited for subsequent attribute-missing imputation tasks. Concretely, we encourage each attribute-incomplete sample to be closer to its counterpart as well as $r$-order neighbors across structure-attribute modalities, which can be formulated as:

\begin{small}
\begin{equation} \label{eq:3}
\mathcal{L}_C = \underbrace{\frac{1}{N^I}\sum_{i}^{N^I}(\mathbf{C}_{ii}-1)^2}_{\text{self-loop consistency}}+ \underbrace{\frac{1}{N^{I}(N^I-1)}\sum_{i}^{N^I}\sum_{j\neq i}^{N^I} (\mathbf{C}_{ij}-\mathbf{\widetilde{A}}^{Ir}_{ij})^2}_{\text{high-order structural consistency}},
\end{equation}
\end{small}
\begin{equation} \label{eq:4}
\mathbf{C} = \frac{\mathbf{H}^{I(1)}_{A}p(\mathbf{H}^{(1)}_{S})^{\rm{T}}}{\Vert \mathbf{H}^{I(1)}_{A} \Vert \Vert p(\mathbf{H}^{(1)}_{S})\Vert},
\end{equation}
where $\mathbf{C} \in \mathbb{R}^{N^{I} \times N^{I}}$ and $\mathbf{\widetilde{A}}^{Ir} \in \mathbb{R}^{N^{I} \times N^{I}}$ denote a structure-attribute correlation matrix and a $r$-order normalized adjacency matrix, respectively. $N^{I}$ is the number of attribute-incomplete samples. In addition, $\mathbf{H}^{I(1)}_{A}$ and $\mathbf{H}^{(1)}_{S}$ indicate the embeddings of attribute-incomplete samples and graph structure in the first layer, respectively. $p(\cdot)$ is an embedding pick-out function.

As seen in Eq.(\ref{eq:3}), the first term makes diagonal elements of the structure-attribute correlation matrix close to one value, causing the structure-attribute embeddings of each attribute-incomplete sample to be consistent. Moreover, the neighbors of each sample contain rich complementary information that should be considered for incomplete attribute completion. Hence, the second term encourages each sample to be closer to $r$-order neighbors than non-neighbors across structure-attribute modalities, aiming to exploit diverse complete structure properties to assist the feature completion of attribute-incomplete samples. By doing this, both sample denoising and structure-attribute consistency constraint are seamlessly integrated to 1) make the learned representations of attribute-incomplete samples robust and invariant to data perturbations (\textit{e.g.,} the noise and incompleteness of the graph); 2) preserve more trustworthy features from available information to achieve better data imputation. The overall pipeline of training the proposed STC is summarized in Algorithm 1.

\begin{algorithm}[!t]
{\caption{The learning procedure of the STC mechanism}\label{Algorithm-proposed}
\small
\begin{algorithmic}[1]
\REQUIRE Sub-graph $\mathcal{G}^{Sub}=\{\mathbf{X}^I, \widetilde{\mathbf{A}}^I\}$; structural graph $\mathcal{G}^S =\{\mathbf{I}, \widetilde{\mathbf{A}}\}$; $r$-order normalized adjacency matrix $\widetilde{\mathbf{A}}^{Ir}$ of $\mathcal{G}^{Sub}$; pre-training iterations $\textit{T}$; hyper-parameters $\beta$. 
\ENSURE Attribute embeddings of incomplete samples $\mathbf{H}^{I}_{A}$ and structure embeddings $\mathbf{H}_S$.
\FOR {$\textit{t} = 1$ to $\textit{T}$}

\STATE Generate a Gaussian noise matrix and assign it to $\mathbf{X}^I$;
\STATE Utilize $E_A(\cdot)$ to learn $\mathbf{H}^{I}_{A}$ by Eq.(\ref{eq:1});
\STATE Utilize $E_S(\cdot)$ to learn $\mathbf{H}_S$ by Eq.(\ref{eq:2});
\STATE Calculate the structure-attribute correlation matrix by Eq.(\ref{eq:4})
\STATE Optimize the model with Adam by minimizing Eq.(\ref{eq:3}) with $\beta$;
\ENDFOR
\RETURN $\mathbf{H}^{I}_{A}$ and $\mathbf{H}_S$
\end{algorithmic}}
\end{algorithm}

\subsection{Initializing Then Refining}
\noindent{\textbf{Imputation Initialization.}}
Besides completing and fine-tuning the samples with incomplete attributes, assigning initial values to the samples with missing samples should be further taken into account. A widely-adopted measure is traditional imputation techniques, such as zero value filling and mean value filling. Nevertheless, in the attribute-missing circumstance, these filling methods could incorporate amounts of irrelevant noise that will diffuse through the network, causing semantically biased representations. To alleviate this issue, it is intuitive to leverage the structure embeddings as the embedding initialization for latent variables of attribute-missing samples. The reason for that is two-fold. Firstly, the attribute embedding and the structure embedding describe different aspects of a node, providing consistent and complementary information in these two modalities \cite{2018DANE}. Secondly, this initialization approach is reliable since the structure information of the original graph is complete. 

To this end, after encoding incomplete attribute embeddings $\mathbf{H}^{I}_{A} \in \mathbb{R}^{N^I \times d}$ and structure embeddings $\mathbf{H}_S \in \mathbb{R}^{N \times d}$, we first pick out the structure embeddings of attribute-missing samples $\mathbf{H}_S^M \in \mathbb{R}^{N^{M} \times d}$ from $\mathbf{H}_S$, and then utilize a Concat function $C(\cdot)$ to integrate $\mathbf{H}_S^M$ with $\mathbf{H}^{I}_{A}$, where $d$ refers to the embedding dimension. It is worth noting that the information concatenation to construct $\mathbf{H}_I \in \mathbb{R}^{N \times d}$ is not the classic channel-wise or row-wise concatenation. In this operation, the latent variables of attribute-incomplete samples are filled with $\mathbf{H}^{I}_{A}$ and the latent variables of attribute-missing samples are filled with $\mathbf{H}_S^M$:
\begin{equation} \label{eq:5}
\mathbf{H}_I = C(\mathbf{H}^{I}_{A}, \mathbf{H}_S^M),
\end{equation}
where $\mathbf{H}_I$ indicates the initially imputed embeddings. In our concatenation settings, the location of each sample remains unchanged within the original graph. 

\noindent{\textbf{Imputation Refinement.}}
It is well known that node attributes preserve the semantic graph content while the graph structure implies the connection relationships among samples. Consequently, the trustworthiness degrees of two-source information exhibit differences to some extent. Making full use of the trustworthy visible structure and attribute information to initial and refine the missing values may benefit the data imputation quality of attribute-missing samples. To illustrate whether our argument holds, we make a comparison among three methods and discuss the performance on Cora and Citeseer. Here we set both attribute-incomplete and attribute-missing ratios as 60\%. Ours-Z, Ours-S, and Ours-S-A are methods where we impute the latent variables of attribute-missing samples with zero values, the structure embeddings merely, and the structure-attribute embeddings, respectively. As shown in Table \ref{II}, we can observe that 1) Ours-S performs better than Ours-Z, which empirically verifies that the structure embeddings $\mathbf{H}_S^M$ could provide an effective embedding initialization; 2) the trustworthy attributes adding in data imputation could provide more discriminative information to assist in the refinement of initial imputation. 
Specifically, we leverage available attribute properties $\mathbf{H}^{I}_{A}$ to refine the initially imputed variables $\mathbf{H}_S^M$ via an affinity structure $\mathbf{R} \in \mathbb{R}^{N \times N}$. The intuition behind this is that $\mathbf{H}^{I}_{A}$ is trustworthy since all attribute-incomplete samples are carefully restored via the STC mechanism. By doing this, the semantic gap between $\mathbf{H}^{I}_{A}$ and $\mathbf{H}_S^M$ is allowed to be narrowed, thus boosting the discriminative capacity of the overall graph embedding.
The above procedure can be written as a graph convolution-like formulation:
\begin{equation} \label{eq:6}
\mathbf{H} = \mathbf{R}\mathbf{H}_I,
\end{equation}
where $\mathbf{H} \in \mathbb{R}^{N \times d}$ indicates the imputed embeddings and we initialize the affinity structure $\mathbf{R}$ as $\widetilde{\mathbf{A}}$. 

\begin{table}[!t]
\centering
\caption{Performance comparison. Both attribute-incomplete and attribute-missing ratios are set to 60\%. $\uparrow$ denotes the performance improvement of Ours-S-A against Ours-S. The boldface and underline values indicate the best and the sub-optimal results (\%), respectively.}
\begin{tabular}{c|c|cc}
\hline
Dataset               & Method  & Recall@50  &   NDCG@50 \\\hline
\multirow{3}{*}{Cora} & Ours-Z & 27.85 & 25.68 \\
                      & Ours-S & \underline{28.59}  & \underline{26.45} \\
                      & Ours-S-A & \textbf{32.43 (3.84 $\uparrow$)}   & \textbf{30.88 (4.43 $\uparrow$)}     \\\hline
\multirow{3}{*}{Citeseer} 
                      & Ours-Z & 18.56 &  18.56 \\
                      & Ours-S  & \underline{19.79}  &    \underline{19.38}   \\
                      & Ours-S-A & \textbf{22.14 (2.35 $\uparrow$)}  & \textbf{23.67 (4.29 $\uparrow$)}     \\
                     \hline

\end{tabular}

\label{II}
\end{table}

According to Eq.(\ref{eq:6}), we refine the attribute-missing imputation from the following two aspects. On the one hand, it is obvious that the noise information in $\mathbf{H}_S^M$ can be transferred into the well-learned attribute embeddings of attribute-incomplete samples. This would undermine the representation quality and the reconstruction accuracy of available information, which in turn negatively affects the subsequent data imputation tasks and even distort the original graph. To tackle this problem, we implement an information recomposing scheme to decrease the adverse effect of noise information passing from the embeddings of attribute-missing samples. Firstly, we pick out the latent variables of attribute-missing samples $\mathbf{H}^M \in \mathbb{R}^{N^M \times d}$ from $\mathbf{H}$, and next recompose them with $\mathbf{H}^{I}_{A}$ using a Concat function $C(\cdot)$:
\begin{equation} \label{eq:7}
\mathbf{\widetilde{H}} = C(\mathbf{H}^{I}_{A}, \mathbf{H}^M),
\end{equation}
where $\mathbf{\widetilde{H}} \in \mathbb{R}^{N \times d}$ indicates the sample-recomposed embeddings. The information recomposing scheme replaces the adjusted embeddings of attribute-incomplete samples with more reliable $\mathbf{H}^{I}_{A}$. Meanwhile, as illustrated in Fig. \ref{2}, we fix the embeddings of attribute-incomplete samples as $\mathbf{H}^{I}_{A}$ in the final step. This provides the most trustworthy information on attribute-incomplete samples for subsequent missing attribute imputation.

\begin{algorithm}[!t]
{\caption{The learning procedure of the ITR mechanism}\label{Algorithm-proposed}
\small
\begin{algorithmic}[1]
\REQUIRE Attribute embeddings of incomplete samples $\mathbf{H}^{I}_{A}$; structure embeddings $\mathbf{H}_S$; iterations $\textit{T}$; update interval $\textit{U}$; element-wise indicator matrix $\mathbf{M}$; hyper-parameters $\gamma$ and $\alpha$.
\ENSURE Rebuilt attribute matrix $\widehat{\mathbf{X}}$ and adjacency matrix $\widehat{\mathbf{A}}$.
\FOR {$\textit{t} = 1$ to $\textit{T}$}
\STATE Obtain $\mathbf{H}_S^M$ from $\mathbf{H}_S$ in accordance with indices;
\STATE Combine $\mathbf{H}^{I}_{A}$ with $\mathbf{H}_S^M$ for initial imputation by Eq.(\ref{eq:5});
\STATE Conduct the imputation refinement using $\mathbf{R}$ by Eq.(\ref{eq:6});
\STATE Recompose the latent variables to obtain $\mathbf{\widetilde{H}}$ by Eq.(\ref{eq:7});
\STATE Calculate $\mathbf{S}$ over the resultant $\mathbf{\widetilde{H}}$ by Eq.(\ref{eq:8});
\IF {$ \textit{t} \% \textit{U} == 0$}
\STATE Update $\mathbf{R}$ to refine $\mathbf{H}$ by Eq.(\ref{eq:9}) and Eq.(\ref{eq:6});
\ENDIF
\STATE Utilize $D(\cdot)$ to decode $\mathbf{\widetilde{H}}$ and output $\widehat{\mathbf{X}}$ by Eq.(\ref{eq:10});
\STATE Utilize a simple inner decoder to rebuild $\widehat{\mathbf{A}}$;
\STATE Optimize the model with Adam by minimizing Eq.(\ref{eq:11}) and Eq.(\ref{eq:12}) with $\alpha$;
\ENDFOR
\RETURN $\widehat{\mathbf{X}}$ and $\widehat{\mathbf{A}}$
\end{algorithmic}}
\end{algorithm}

\begin{table*}[!t]
\caption{A comparison of the complexity analysis among SAT \cite{2020SAT}, ITR \cite{2022ITR}, SVGA \cite{2022SVGA}, and the proposed RITR.}
\centering
\begin{tabular}{c|c}\hline
Model   & Time complexity (model and loss error computation)    \\\hline
SAT    &   $\mathcal{O}(Nd(D_{a}+D_{s}+dL)+|\mathcal{E}|dL+N^2+nD_a)$ $\approx$ $\mathcal{O}(N^2)$     \\
SVGA      &  $\mathcal{O}(Nd^2L+|\mathcal{E}|dL+NdD_{a}+Ndc+ND_a+N+N^2)$ $\approx$ $\mathcal{O}(N^2)$     \\
ITR   &   $\mathcal{O}(Nd(D_{a}+D_{s}+dL)+|\mathcal{E}|(dL+D_{a})+N^2+nD_a)$ $\approx$ $\mathcal{O}(N^2)$      \\\hline
RITR (Ours)  &  $\mathcal{O}(Nd(D_{a}+D_{s}+dL)+|\mathcal{E}|(dL+D_{a})+N^2+n(n+D_a))$ $\approx$ $\mathcal{O}(N^2)$       \\
\hline
\end{tabular}
\label{III}
\end{table*}

On the other hand, we argue that the initial affinity matrix $\textbf{R}$ (\textit{i.e.,} $\widetilde{\textbf{A}}$) is not the ground truth. The limitations within this matrix are two-fold: 1) noisy connections. Besides inner connections within clusters, inappropriate connections could exist between clusters in the matrix; 2) missing connections. In $\widetilde{\textbf{A}}$, only the first-order connections are preserved, the high-order relevant connections could be missing. Both would cause inaccurate imputation and reconstruction of missing attributes.
To overcome these issues, we seek to refine $\textbf{R}$ by emphasizing the dependable connections while weakening the unreliable ones. To this end, we propose an affinity structure updating scheme to optimize $\textbf{R}$ with iterations.  
Specifically, we first calculate a normalized self-correlated matrix $\mathbf{S} \in \mathbb{R}^{N \times N}$ according to $\mathbf{\widetilde{H}}$ as below:
\begin{equation} \label{eq:8}
\mathbf{S}_{jk} = \mathcal{N}(\frac{{\mathbf{\widetilde{h}}_{j}\mathbf{\widetilde{h}}_{k}^{\rm{T}}}}{{\Vert \mathbf{\widetilde{h}}_{j} \Vert \Vert \mathbf{\widetilde{h}}_{k} \Vert}}), \,\,\,\,  \forall\,\,j, k \in [1, N],
\end{equation}
where $\mathcal{N}(\mathbf{Y}) = \mathbf{D}_{\mathbf{Y}}^{-\frac{1}{2}}\mathbf{Y} \mathbf{D}_{\mathbf{Y}}^{-\frac{1}{2}}$ indicates a structural normalization function, $\mathbf{D}_{\mathbf{Y}} \in \mathbb{R}^{N \times N}$ is the degree matrix of $\mathbf{Y}$. $\mathbf{\widetilde{h}}_{j}$ ($\mathbf{\widetilde{h}}_{k}$) indicates the embedding of sample $\mathbf{v}_j$ ($\mathbf{v}_k$).
Then we optimize the affinity structure $\mathbf{R}$ every $\textit{t}$ iterations via Eq.(\ref{eq:9}), and leverage it as guidance for the subsequent missing attribute imputation: 
\begin{equation} \label{eq:9}
\mathbf{R} = \gamma\widetilde{\mathbf{A}}+(1-\gamma)\mathbf{S},
\end{equation}
where $\gamma$ is a balanced hyper-parameter and is initialized as 0.5. 
With the affinity structure updating scheme, the network is enabled to construct the embeddings of attribute-missing samples with not only the first-order but also the high-order connections within the graph structure. Since the embeddings of attribute-incomplete samples become more reliable and the embeddings of attribute-missing samples become more informative, the quality of data imputation could be further improved, making the learned representations more discriminative and robust. The overall pipeline of training the proposed ITR is summarized in Algorithm 2.

\subsection{Training Objectives and Complexity Analysis}
\subsubsection{Training Objectives}
After obtaining $\mathbf{\widetilde{H}}$, we feed it with $\widetilde{\mathbf{A}}$ into a graph decoder $D(\cdot)$ to rebuilt the attributes of attribute-incomplete and attribute-missing samples:
\begin{equation} \label{eq:10}
\mathbf{\widetilde{H}}^{(l)} = \sigma(\widetilde{\mathbf{A}}\mathbf{\widetilde{H}}^{(l-1)}\mathbf{\Phi}^{(l)}),
\end{equation}
where $\mathbf{\widetilde{W}}^{(l)}$ indicates the parameter matrix of $D(\cdot)$ in the \textit{l}-th layer. $\mathbf{\widetilde{H}}^{(0)}$ and $\mathbf{\widetilde{H}}^{(2)}$ denote the sample-recomposed embeddings $\mathbf{\widetilde{H}}$ and the rebuilt attribute matrix $\widehat{\mathbf{X}} \in \mathbb{R}^{N \times D}$, respectively.
The joint loss function of RITR includes three parts, which can be written as:
\begin{equation} \label{eq:11}
\mathcal{L}_{A} = \frac{1}{2N^I}\Arrowvert\mathbf{M}\odot(\mathbf{X}^I - \widehat{\mathbf{X}}^I)\Arrowvert_F^2,
\end{equation}

\begin{equation} \label{eq:12}
\mathcal{L}_{S} = \frac{1}{N^{2}}\sum_{i=1}^N \sum_{j=1}^N {BCE(\widetilde{\mathbf{A}}_{ij}}, \widehat{\mathbf{A}}_{ij}),
\end{equation}
\begin{equation} \label{eq:13}
\mathcal{L} = \alpha \mathcal{L}_{A} + \mathcal{L}_{S} + \beta \mathcal{L}_{C}.
\end{equation}
In Eq.(\ref{eq:11}), $\mathcal{L}_{A}$ refers to the mean square error (MSE) of attribute-incomplete samples between $\mathbf{X}$ and $\mathbf{\widehat{X}}$. $\mathbf{M} \in \mathbb{R}^{N^{I} \times D}$ is an element-wise indicator matrix where $\mathbf{M}_{ij}$ = 1 if $\mathbf{X}^I_{ij}$ is a real value, otherwise $\mathbf{X}^I_{ij}$ is a null value (\textit{i.e.,} an incomplete attribute). 
In Eq.(\ref{eq:12}), $\mathcal{L}_{S}$ refers to the binary cross-entropy (BCE) between the normalized adjacency matrix $\widetilde{\mathbf{A}}$ and the rebuilt adjacency matrix $\widehat{\mathbf{A}} \in \mathbb{R}^{N \times N}$, where $\widehat{\textbf{A}}=\sigma(\textbf{H}_S \widetilde{\textbf{H}}^{\rm{T}})$, $\sigma(\cdot)$ is a Sigmoid activation function. $\alpha$ and $\beta$ are two balanced hyper-parameters. The applied optimization objectives are similar to existing attribute-missing graph machine learning methods \cite{2020SAT, 2022ITR}. However, the major differences between current methods and our improved method could be summarized in the following three parts: 1) more naturally handling hybrid-absent graphs in an unsupervised circumstance; 2) more comprehensive that seamlessly unifies the representation learning and data imputation processes of attribute-incomplete and attribute-missing samples into a common framework; 3) more discriminative that enables the structure-attribute information to sufficiently negotiate with each other for feature completion by performing STC and ITR mechanisms.

\subsubsection{Complexity Analysis}
The time complexity of the proposed RITR could be discussed from the following two aspects: the graph auto-encoder framework and the loss error computation. For two GCN-based graph encoders, the complexities of $E_A(\cdot)$ and $E_s(\cdot)$ are $\mathcal{O}(Nd^2(L-1)+NdD_{a}+|\mathcal{E}|dL)$ and $\mathcal{O}(Nd^2(L-1)+NdD_{s}+|\mathcal{E}|dL)$, where $N$, $L$, $|\mathcal{E}|$ are the number of nodes, encoder layers, and edges, respectively. $D_{a}$, $D_{s}$, and $d$ are the dimension of raw attribute features, raw structure features, and hidden representations, respectively. For the graph decoder, the complexity of $D(\cdot)$ is $\mathcal{O}(Nd^2(L-1)+NdD_{a}+|\mathcal{E}|d(L-1)+|\mathcal{E}|D_{a})$. For the loss error computation, we follow SAT \cite{2020SAT} and use the MSE and BCE loss functions to reconstruct node attributes and graph structure, respectively. The time complexities of $\mathcal{L}_A$ and $\mathcal{L}_S$ are $\mathcal{O}(nD_a)$ and $\mathcal{O}(N^2)$. The time complexity of the structure-attribute consistency loss function $\mathcal{L}_C$ is $\mathcal{O}(n^2)$, where $n=(1-\theta)N$, $\theta$ is the ratio of attribute-missing samples. Considering the attribute-missing problem is universal in real-world scenarios (\textit{i.e.,} $\theta$ can be set to a large value) and $n^2$ can be a relatively small value, the computation overhead here is acceptable. The overall time complexity of RITR for each training iteration is $\mathcal{O}(Nd(D_{a}+D_{s}+dL)+|\mathcal{E}|(dL+D_{a})+N^2+n(n+D_a))$ $\approx$ $\mathcal{O}(N^2)$. For a fair comparison, we conduct a complexity comparison among four attribute-missing graph machine learning methods and report the results in Table \ref{III}. As seen, RITR consistently outperforms all baselines in both attribute-missing and hybrid-absent cases (see Sections V-B and V-C), requiring no additional computation complexity compared to these competitors. 

\section{Experiments}
In this section, we evaluate the effectiveness of ITR and RITR against some advanced graph machine learning methods. The experiments aim to answer the following five questions:
\begin{itemize}
\item $\mathbf{Q1.}$ How do the proposed methods perform compared to baselines in profiling and node classification tasks? 
\item $\mathbf{Q2.}$ How does the designed component influence the performance? 
\item $\mathbf{Q3.}$ How does the proposed method perform with different absent ratios?
\item $\mathbf{Q4.}$ How do key hyper-parameters influence the performance of the proposed method?  
\item $\mathbf{Q5.}$ How about the method convergence and performance variation with iterations?
\end{itemize}
In the following, we begin with a brief experimental setup introduction, including benchmark datasets, implementation procedures, training settings, and compared methods. Then we report experimental results with corresponding analysis.

\subsection{Experimental Setup}
\subsubsection{Benchmark Datasets} We implement experiments to evaluate two proposed methods, \textit{i.e.,} ITR and RITR on four benchmark datasets, including Cora, Citeseer, Amazon Computer, and Amazon Photo. We summarize the detailed dataset information in Table \ref{IV}. 

\begin{itemize}
\item Cora$\footnote{https://docs.dgl.ai/api/python/dgl.data.html\#citation-network-dataset\label{citation}}$ and Citeseer$\textsuperscript{\ref {citation}}$ are two popular citation network datasets. Specially, nodes mean scientific publications, and edges mean citation relationships. Each node has a predefined feature with corresponding dimensions.
\item Amazon Photo$\footnote{https://docs.dgl.ai/api/python/dgl.data.html\#amazon-co-purchase-dataset\label{amazon}}$ and Amazon Computers$\textsuperscript{\ref {amazon}}$ (Amap and Amac for abbreviation) are segments of the Amazon co-purchase network, where nodes represent goods, edges indicate that two goods are frequently bought together, node features are bag-of-words encoded product reviews, and class labels are given by the product category.
\end{itemize}


\begin{table}[!t]
\centering
\caption{Summary of datasets.}
\begin{tabular}{c|c|c|c|c}\hline
Dataset    & Nodes   & Edges    & Dimension  & Classes\\\hline
Cora       & 2708    & 5278    & 1433       & 7       \\
Citeseer   & 3327    & 4228     & 3703       & 6       \\
Amac   & 13752   & 245861   & 767        & 10      \\
Amap   & 7650    & 119081   & 745        & 8      \\
\hline
\end{tabular}
\label{IV}
\end{table}

\begin{table*}[!t]
\centering
\caption{Profiling performance comparison in the attribute-missing circumstance. The attribute-missing ratio is set to 60\%. The boldface and underline values indicate the best and the sub-optimal results (\%), respectively. }
\scriptsize
\begin{tabular}{c|c|cccccccccc|cc}
\hline
Dataset               & Metric & NeighAggre & VAE    & GCN   & GraphSage  & GAT   & Hers & GraphRNA & ARWMF & SVGA&SAT  & ITR & RITR
\\\hline
\multirow{6}{*}{Cora} & Recall@10    & 9.06  & 8.87  & 12.71  & 12.84  & 13.50  & 12.26  & 13.95  & 12.91  & 16.06 &15.08  & \underline{16.82}  &  \textbf{17.32}\\
                      & Recall@20    & 14.13  & 12.28  & 17.72  & 17.84  & 18.12  & 17.23  & 20.43  & 18.13  & 23.17 & 21.82  & \underline{23.69}  & \textbf{24.47}\\
                      & Recall@50    & 19.61  & 21.16  & 29.62  & 29.72  & 29.72  & 27.99  & 31.42  & 29.60 & 35.83 & 34.29 &  \underline{36.47}  & \textbf{36.81}\\
                      & NDCG@10      & 12.17  & 12.24  & 17.36  & 17.68  & 17.91  & 16.94  & 19.34  & 18.24  &23.19  & 21.12 & \underline{23.20}  & \textbf{24.20}\\
                      & NDCG@20      & 15.48  & 14.52  & 20.76  & 21.02  & 20.99  & 20.31  & 23.62  & 21.82   & 26.81 & 25.46 &\underline{27.81}  & \textbf{28.97}\\
                      & NDCG@50      & 18.50  & 19.24  & 27.02  & 27.28  & 27.11  & 25.96  & 29.38  & 27.76  & 33.71 & 32.12  &\underline{34.60}  & \textbf{35.52}\\\hline
\multirow{6}{*}{Citeseer} 
                      & Recall@10    & 5.11 & 3.82 & 6.20 & 6.12 & 5.61 & 5.76 & 7.77 & 5.52 & 8.83 & 7.64&  \underline{9.63} & \textbf{10.16}\\
                      & Recall@20    & 9.08 & 6.68 & 10.97 & 10.97 & 10.12 & 10.25 & 12.72 & 10.15 & 14.55 & 12.80 & \underline{15.48}  & \textbf{15.97}\\
                      & Recall@50    & 15.01 & 12.96 & 20.52 & 20.58 & 19.57 & 19.73 & 22.71 & 19.52& 25.91 & 23.77 & \underline{26.84} & \textbf{26.90}\\
                      & NDCG@10      & 8.23 & 6.01 & 10.26 & 10.03 & 8.78 & 9.04 & 12.91 & 8.59 &15.21& 12.98 & \underline{16.32} & \textbf{17.16}\\
                      & NDCG@20      & 11.55 & 8.39 & 14.23 & 13.93 & 12.53 & 12.79 & 17.03 & 12.45  & 20.08 & 17.29& \underline{21.20} & \textbf{22.03}\\
                      & NDCG@50      & 15.60 & 12.51 & 20.49 & 20.34 & 18.72 & 19.00 & 23.58 & 18.58  & 27.99 & 24.47& \underline{28.69} & \textbf{29.22}\\\hline
\multirow{6}{*}{Amac}
                      & Recall@10    & 3.21 & 2.55 & 2.73 & 2.69 & 2.71 & 2.73 & 3.86 & 2.80 & 3.97 &3.91 & \underline{4.48} & \textbf{4.59}\\
                      & Recall@20    & 5.93 & 5.02 & 5.33 & 5.28 & 5.30 & 5.25 & 6.90 & 5.44 & 7.22& 7.03 &\underline{7.81} & \textbf{8.06}\\
                      & Recall@50    & 13.06 & 11.96 & 12.75 & 12.78 & 12.78 & 12.73 & 14.65 & 12.89 & 15.75 &15.14 & \underline{16.20} & \textbf{16.48}\\
                      & NDCG@10      & 7.88 & 6.32 & 6.71 & 6.64 & 6.73 & 6.76 & 9.31 & 6.94 & 10.11 &9.63 & \underline{10.95} & \textbf{11.14}\\
                      & NDCG@20      & 11.56 & 9.70 & 10.27 & 10.20 & 10.28 & 10.25 & 13.33 & 10.53 & 14.98 &13.79 & \underline{15.37} & \textbf{15.69}\\
                      & NDCG@50      & 19.23 & 17.21 & 18.24 & 18.22 & 18.30 & 18.25 & 21.55 & 18.51 & 23.84&22.43 & \underline{24.29} & \textbf{24.63}\\\hline
\multirow{6}{*}{Amap}  
                      & Recall@10    & 3.29 & 2.76 & 2.94 & 2.95 & 294 & 2.92 & 3.90 & 2.94 & 4.14 & 4.10  &\underline{4.36} &\textbf{4.49}\\
                      & Recall@20    & 6.16 & 5.38 & 5.73 & 5.62 & 5.73 & 5.74 & 7.03 & 5.68 & 7.51 &7.43 & \underline{7.82} &\textbf{8.00}\\
                      & Recall@50    & 13.61 & 12.79 & 13.24 & 13.22 & 13.24 & 13.28 & 15.08 & 13.27 & 15.93 &15.97 & \underline{16.43} &\textbf{16.61}\\
                      & NDCG@10      & 8.13 & 6.75 & 7.05 & 7.12 & 7.05 & 7.14 & 9.59 & 7.27 & 10.21 &10.06 & \underline{10.73} &\textbf{10.94}\\
                      & NDCG@20      & 11.96 & 10.31 & 10.82 & 10.79 & 10.83 & 10.94 & 13.77 & 10.98 & 14.73 &14.50 & \underline{15.33} &\textbf{15.56}\\
                      & NDCG@50      & 19.98 & 18.30 & 18.93 & 18.96 & 18.92 & 19.06 & 22.32 & 19.15 & 23.75 &23.59& \underline{24.50} &\textbf{24.72}\\\hline
\end{tabular}
\label{V}
\end{table*}

\subsubsection{Implementation Procedures}
Both ITR and RITR are implemented with the Pytorch platform. We evaluate the effectiveness of two proposed methods through a two-step learning procedure. Firstly, we train an unsupervised framework to learn node representations and complete absent information for at least 600 iterations. Following SAT \cite{2020SAT}, we regard the profiling learning as a pretext task and adopt Recall@K and NDCG@K as metrics to evaluate the quality of rebuilt attributes. To alleviate the over-fitting problem, we perform an early stop strategy when the loss value reaches a plateau. Secondly, for the node classification task, we feed the rebuilt attribute matrix into a graph classifier and optimize it with five-fold validation 10 times, and report the average accuracy (ACC) performance. 

\subsubsection{Training Settings}
In the attribute-missing case, we record the performance of all methods directly according to the paper of SAT \cite{2020SAT} except for GINN \cite{2020GINN}, GCNMF \cite{2021GCNMF}, and SVGA \cite{2022SVGA}. In the hybrid-absent case, we run the released source code of all compared methods by following the settings of the corresponding literature, and report their results. For our proposed ITR and RITR, we strictly follow the criterion of data splits as was done in SAT, including the split ratio of attribute-complete/missing samples and the split ratio of train/test sets. Specifically, 1) in the profiling task, we randomly sample 40\% nodes with complete attributes as the training set, and manually mask all attributes of the rest of 10\% and 50\% nodes (\textit{i.e.,} attribute-missing samples) as the validation set and the test set, respectively. Besides, when attribute-incomplete and attribute-missing samples exist simultaneously within a graph, we randomly mask 60\% attributes of each attribute-complete sample (\textit{i.e.}, the training set) before network learning. We employ a 4-layer graph auto-encoder framework and optimize it with Adam optimization algorithm. During the training phase, we transfer all samples into ITR and RITR to complete absent attributes by merely reconstructing the available ones. After training, we rebuild the attribute matrix over the well-trained model via forwarding propagation; 2) in the node classification task, we randomly split the rebuilt attributes into 80\% and 20\% for training and testing, respectively. We train the classifier with five-fold validation for 1000 iterations and repeat the experiments 10 times. According to
the results of parameter sensitivity testing, we fix two balanced
hyper-parameters $\alpha$ and $\beta$ to 10. Moreover, the learning rate, the latent dimension, the dropout rate, and the weight decay are set to 1e-3, 64, 0.5, and 5e-4, respectively. Please note that we do not carefully tune these parameters for ease of training as was done in SAT.

\subsubsection{Compared Methods}
We compare RITR with the existing 13 baseline methods for future estimation on attribute-missing and hybrid-absent graphs. Specifically, \textbf{NeighAggre} (NAS’ 08) \cite{2008Navigating} is a classical profiling method. \textbf{VAE} (NeurIPS’ 16) \cite{2014VAE} is a well-known auto-encoder method. \textbf{GCN} (ICLR’ 17) \cite{2017GCN}, \textbf{GraphSage} (NeurIPS’ 17) \cite{2017Graphsage}, and \textbf{GAT} (ICLR’ 18) \cite{2018GAN} are three typical graph neural networks. \textbf{GraphRNA} (KDD’ 19) \cite{2019GraphRNA} and \textbf{ARWMF} (NeurIPS’ 19) \cite{2019ARWMF} are representatives of attributed random walk-based methods. \textbf{Hers} (AAAI’ 19) \cite{2019HERS} is a cold-start recommendation method. \textbf{SAT} (TPAMI’ 22) \cite{2020SAT} is the first attribute-missing graph imputation network. \textbf{SVGA} (KDD' 22) \cite{2022SVGA} and \textbf{ITR} (IJCAI' 22) \cite{2022ITR} are two most advanced attribute-missing graph autoencoders. \textbf{GINN} (NN’ 20) \cite{2020GINN} and \textbf{GCNMF} \cite{2020GINN} (FGCS’ 21) \cite{2021GCNMF} are two state-of-the-art attribute-incomplete graph machine learning methods.

\begin{table*}[!t]
\centering
\caption{Node classification performance comparison. ``AM" refers to the attribute-missing circumstance. ``HA" refers to the hybrid-absent circumstance. Both attribute-incomplete and attribute-missing ratios are set to 60\%. The boldface and underline values indicate the best and the sub-optimal results (\%), respectively.}
\scriptsize
\begin{tabular}{c|c|cc|cc|cc|cc|cc|cc|cc|cc}
\hline

&  &  \multicolumn{2}{c|}{GCN}  & \multicolumn{2}{c|}{GAT}  & \multicolumn{2}{c|}{GINN} & \multicolumn{2}{c|}{GCNMF}  & \multicolumn{2}{c|}{SVGA} & \multicolumn{2}{c|}{SAT}      &\multicolumn{2}{c|}{ITR}          & \multicolumn{2}{c}{RITR}           \\ \cline{3-18} 

\multirow{-2}{*}{{Type}} & \multirow{-2}{*}{{Dataset}} & \multicolumn{1}{c|}{AM}      &  HA    & \multicolumn{1}{c|}{AM}      &    HA   & \multicolumn{1}{c|}{AM}      &    HA   & \multicolumn{1}{c|}{AM}           &    HA        & \multicolumn{1}{c|}{AM}           &     HA       & \multicolumn{1}{c|}{AM}      & HA & \multicolumn{1}{c|}{AM} & HA & \multicolumn{1}{c|}{AM} & HA \\ \hline

& Cora  & \multicolumn{1}{c|}{39.43} & 29.39 & \multicolumn{1}{c|}{41.43} & 29.39 & \multicolumn{1}{c|}{-}          & -          & \multicolumn{1}{c|}{-}          & -         & \multicolumn{1}{c|}{78.70}    &  67.91    & \multicolumn{1}{c|}{76.44} & 72.53     & \multicolumn{1}{c|}{\underline{81.43}}     & \underline{75.04}     & \multicolumn{1}{l|}{\textbf{81.64}} & \textbf{78.11}    \\ 

 & Citeseer   & \multicolumn{1}{c|}{37.68} & 20.96 & \multicolumn{1}{c|}{21.29} & 21.26 & \multicolumn{1}{c|}{-}          & -          & \multicolumn{1}{c|}{-}          & -       & \multicolumn{1}{c|}{62.33}    &   54.51  & \multicolumn{1}{c|}{60.10} & 54.15       & \multicolumn{1}{c|}{\underline{67.15}}     & \underline{61.05}     & \multicolumn{1}{l|}{\textbf{67.47}} & \textbf{63.75}   \\
 
 & Amac     & \multicolumn{1}{c|}{36.60} & 35.76 & \multicolumn{1}{c|}{37.47} & 36.98 & \multicolumn{1}{c|}{-}          & -          & \multicolumn{1}{c|}{-}          & -        & \multicolumn{1}{c|}{ 72.56}    &64.44  & \multicolumn{1}{c|}{74.10} & 68.94       & \multicolumn{1}{c|}{\underline{83.88}}     & \underline{80.68}     & \multicolumn{1}{l|}{\textbf{85.28}} & \textbf{82.92}                 \\ 
 
\multirow{-4}{*}{X}    & Amap &\multicolumn{1}{c|}{26.83} & 25.70 & \multicolumn{1}{c|}{25.98} & 25.39 & \multicolumn{1}{c|}{-} & - & \multicolumn{1}{c|}{-} & -  &  \multicolumn{1}{c|}{ 88.55}    & 82.49& \multicolumn{1}{c|}{87.62} & 84.53         & \multicolumn{1}{c|}{\underline{90.75}}     & \underline{88.98}     & \multicolumn{1}{l|}{\textbf{91.27}} & \textbf{90.49}                 \\ \hline
   & Cora     &  \multicolumn{1}{c|}{43.87} & 30.40 & \multicolumn{1}{c|}{45.25} & 30.80 & \multicolumn{1}{c|}{67.58}      & 38.16      & \multicolumn{1}{c|}{70.30}      &  57.19      & \multicolumn{1}{c|}{83.78}  & 77.70& \multicolumn{1}{c|}{83.27} & 80.37          & \multicolumn{1}{c|}{\underline{85.56}}     & \underline{82.53}     & \multicolumn{1}{c|}{\textbf{85.81}} & \textbf{84.29}                 \\ 
   & Citeseer   & \multicolumn{1}{c|}{40.79} & 26.89 & \multicolumn{1}{c|}{26.88} & 26.49 & \multicolumn{1}{c|}{55.32}      &   23.66         & \multicolumn{1}{c|}{63.40}      &  50.90         &\multicolumn{1}{c|}{ 66.19}    & 60.30  & \multicolumn{1}{c|}{65.99} & 64.16       &  \multicolumn{1}{c|}{\underline{68.09}}     & \underline{64.93}     & \multicolumn{1}{l|}{\textbf{69.01}} & \textbf{66.00}                 \\ 
   & Amac  &\multicolumn{1}{c|}{39.74} & 38.88 & \multicolumn{1}{c|}{40.34} & 39.78 & \multicolumn{1}{c|}{81.27}      &      25.45      & \multicolumn{1}{c|}{76.43}      &   35.40      &  \multicolumn{1}{c|}{85.87} & 80.56  & \multicolumn{1}{c|}{85.19} & 83.33        & \multicolumn{1}{c|}{\underline{87.65}}     & \underline{84.96}     & \multicolumn{1}{l|}{\textbf{88.49}} & \textbf{86.19}                 \\ 
\multirow{-4}{*}{X+A}   & Amap  & \multicolumn{1}{c|}{36.56} & 36.08 & \multicolumn{1}{c|}{37.89} & 37.75 & \multicolumn{1}{c|}{87.77}      &       37.25     & \multicolumn{1}{c|}{87.79}      &       82.04     & \multicolumn{1}{c|}{89.90}    & 84.33 & \multicolumn{1}{c|}{91.63} & 90.49         & \multicolumn{1}{c|}{\underline{91.87}}     & \underline{88.98}     & \multicolumn{1}{l|}{\textbf{92.24}} & \textbf{91.75}                 \\ \hline
\end{tabular} 
\label{VI}
\end{table*}

\subsection{Attribute-missing Case}
\subsubsection{Performance Comparison (Q1)}
As shown in Table \ref{V}, we report the profiling performance of all methods mentioned above. This table shows that ITR and RITR outperform all compared baseline methods in terms of six metrics on four datasets. Specifically, 1) we first compare NeighAggre and VAE with our methods. Instead of merely exploiting the structure or attribute information for data imputation, our methods have two-source information sufficiently negotiate with each other, thus consistently exceeding NeighAggre and VAE by a large margin; 2) ITR and RITR show superior performance against GCN, GraphSage, and GAT, all of which have demonstrated strong representation learning capability in handling attribute-complete graphs. However, the results show that these methods are not suitable to solve the attribute-missing problem; 3) for the two strongest attribute-missing graph machine learning methods (\textit{i.e.,} SVGA and SAT), RITR outperforms them by 1.81\%/3.40\%, 1.23\%/4.75\%, 0.79\%/2.20\%, and 0.97\%/1.13\% in terms of NDCG@50 metric on four datasets, respectively. This is because these baselines heavily rely on pre-defined assumptions that may not always hold in real-world graphs for data imputation, while RITR does not make any prior distribution assumption so that it can flexibly and effectively make full use of visible information for feature completion; 4) RITR further achieves better performance than ITR on all datasets. These superior results of RITR over the state-of-the-art method further verify the effectiveness of our improved framework for handling attribute-missing graphs.

Moreover, we report the node classification performance of 8 methods in Table \ref{VI}. ``X" or ``X+A" indicates that the classifier receives the attribute matrix or attribute and adjacency matrices as input in the node classification task. Note that here we only take the attribute-missing case (\textit{i.e.,} marked as ``AM") into consideration. From these results, we can see that 1) the classification results of GINN and GCNMF are not comparable to those of our two methods. ITR and RITR achieve at least 15.26\%/15.51\%, 4.69\%/5.61\%, 6.38\%/7.22\%, and 4.08\%/4.45 accuracy increment. This indicates that these attribute-incomplete methods fall into inaccurate data imputation with extremely limited observations so that they can not learn effective representations; 2) taking the performance of ``X" for instance, ITR and RITR gain 4.99\%/5.20\%, 7.05\%/7.37\%, 9.78\%/11.18\%, and 3.13\%/3.65\% performance enhancement over the state-of-the-art SAT method. Similar observations can be obtained among SVGA, ITR, and RITR. These benefits can be attributed to the following merits: 1) different from SVGA and SAT, our proposed graph imputation networks avoid the reliance on any prior distribution assumption for missing attribute completion, so that they can facilitate the structure-attribute negotiation more flexibly and comprehensively; 2) the trustworthy visible attribute information and structure information can be used unitedly by ITR and RITR for data imputation instead of being treated separately. The above experimental results well demonstrate the superiority of ITR and RITR in the attribute-missing case.

\begin{figure}[!t]
		\centering
		\includegraphics[width=3.15in]{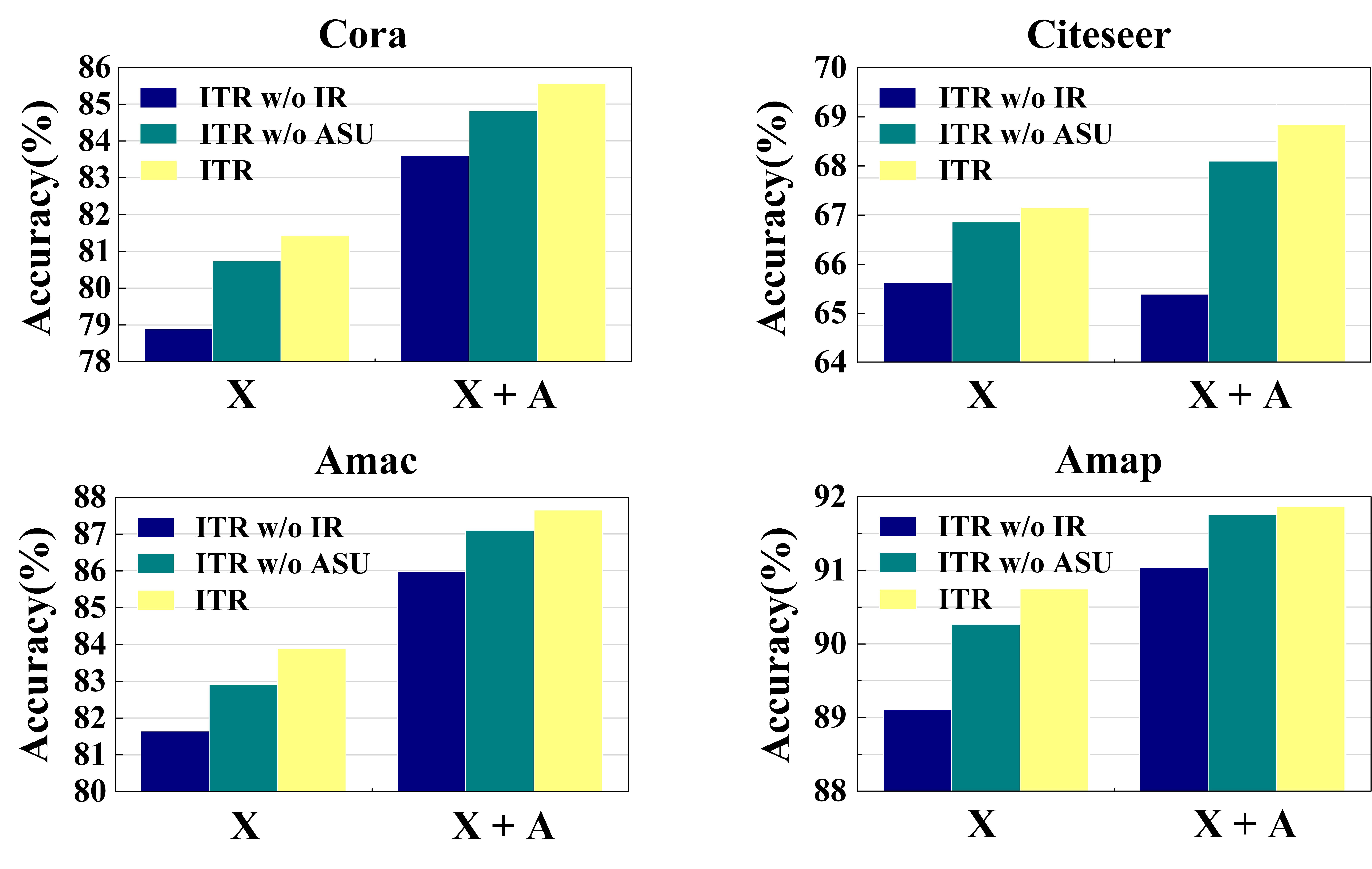}
		\caption{Effect of the information recomposing (IR) and affinity structure updating (ASU) schemes for node classification.}
		\label{3}
\end{figure}

\subsubsection{Effect of Two Schemes in ITR Criterion (Q2)}
\begin{figure}[!t]
		\centering
		\includegraphics[width=3.065in]{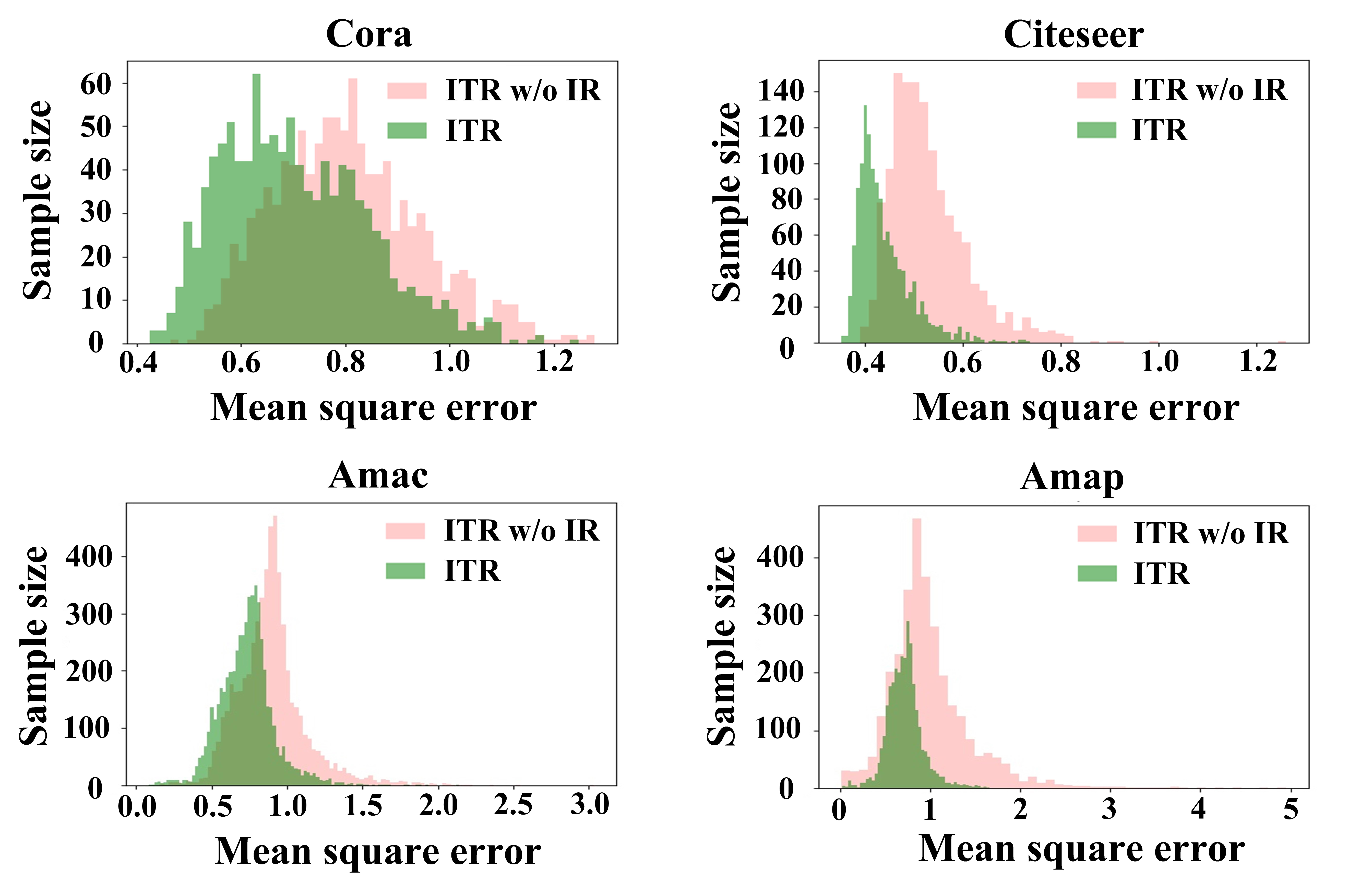}
		\caption{MSE comparison of ITR and an ITR variant. ITR w/o IR denotes the ITR without information recomposing scheme.}
		\label{4}
\end{figure}

To verify the benefit of the initializing-then-refining imputation criterion, we conduct ablation studies on four datasets to compare ITR and two ITR variants, each of which has one of the critical components removed. ITR w/o IR and ITR w/o ASU indicate the method with information recomposing and affinity structure updating being masked, respectively. From the results in Fig. \ref{3}, we observe that the accuracy of ITR on four datasets would degrade without one of the key components. Specifically, for the ``X" task, ITR exceeds ITR w/o IR by 2.54\%, 1.53\%, 2.24\%, and 1.64\% accuracy increment, and ITR w/o ASU by 0.69\%, 0.30\%, 0.98\%, and 0.55\% accuracy increment on Cora, Citeseer, Amac, and Amap, respectively. We find that the information recomposing scheme plays a more important role than the information refining scheme. To visually illustrate this point, we present the mean square error comparison of ITR and ITR w/o IR at the last training iteration. As seen in Fig. \ref{4}, the method with the IR scheme achieves a better convergence than ITR w/o IR. This indicates that our information recomposing operation can effectively prohibit inaccurate information from being propagated, so the network can learn reliable representations for high-quality missing attribute restoration. All the above observations demonstrate the effectiveness of our proposed ITR imputation criterion, which can enable the structure-attribute information to sufficiently negotiate with each other for more accurate data imputation.

\begin{table*}[!t]
\centering
\caption{Ablation study on the effectiveness of each component of RITR. RITR w/o STC and RITR w/o ITR indicate that the method without the sample-denoising then consistency-preserving component and the initializing then refining component, respectively. Both attribute-incomplete and attribute-missing ratios are set to 60\%. The boldface and underline values indicate the best and the sub-optimal results (\%), respectively.}
\begin{tabular}{c|c|cccccc}
\hline
\multicolumn{1}{c|}{Dataset}                     & \multicolumn{1}{c|}{Method}       & \multicolumn{1}{c|}{Recall@10}      & \multicolumn{1}{c|}{Recall@20}      & \multicolumn{1}{c|}{Recall@50}      & \multicolumn{1}{c|}{NDCG@10}        & \multicolumn{1}{c|}{NDCG@20}        & NDCG@50        \\ \hline
{  }                       & {  RITR w/o STC}     & \multicolumn{1}{c|}{\underline{13.63}}          & \multicolumn{1}{c|}{\underline{19.67}}          & \multicolumn{1}{c|}{\underline{30.91}}          & \multicolumn{1}{c|}{\underline{19.45}}          & \multicolumn{1}{c|}{\underline{23.47}}          & \underline{29.43}          \\
                                              & RITR w/o ITR                            & \multicolumn{1}{c|}{12.03}           & \multicolumn{1}{c|}{16.52}          & \multicolumn{1}{c|}{27.70}          & \multicolumn{1}{c|}{16.79}          & \multicolumn{1}{c|}{19.84}          &    25.66        \\   
\multirow{-3}{*}{{  Cora}} & {  RITR} & \multicolumn{1}{c|}{\textbf{14.36}} & \multicolumn{1}{c|}{\textbf{20.69}} & \multicolumn{1}{c|}{\textbf{32.43}} & \multicolumn{1}{c|}{\textbf{20.51}} & \multicolumn{1}{c|}{\textbf{24.69}} & \textbf{30.88} \\ \hline
                                              & RITR w/o STC                            & \multicolumn{1}{c|}{\underline{7.08}}           & \multicolumn{1}{c|}{\underline{11.93}}          & \multicolumn{1}{c|}{\underline{21.61}}          & \multicolumn{1}{c|}{\underline{12.12}}          & \multicolumn{1}{c|}{\underline{16.37}}          & \underline{22.80}          \\
                                              & RITR w/o ITR                            & \multicolumn{1}{c|}{5.19}           & \multicolumn{1}{c|}{9.30}          & \multicolumn{1}{c|}{ 18.17}          & \multicolumn{1}{c|}{8.03}          & \multicolumn{1}{c|}{11.46}          & 17.25 \\     
\multirow{-3}{*}{Citeseer}                    & RITR                        & \multicolumn{1}{c|}{\textbf{7.78}}  & \multicolumn{1}{c|}{\textbf{12.51}} & \multicolumn{1}{c|}{\textbf{22.14}} & \multicolumn{1}{c|}{\textbf{13.38}} & \multicolumn{1}{c|}{\textbf{17.33}} & \textbf{23.67} \\ \hline
                                              & RITR w/o STC                            & \multicolumn{1}{c|}{\underline{3.92}}           & \multicolumn{1}{c|}{\underline{6.97}}           & \multicolumn{1}{c|}{\underline{15.10}}          & \multicolumn{1}{c|}{\underline{9.84}}           & \multicolumn{1}{c|}{\underline{14.05}}          & \underline{22.72}          \\ 
                                              & RITR w/o ITR                            & \multicolumn{1}{c|}{2.83}           & \multicolumn{1}{c|}{5.49}          & \multicolumn{1}{c|}{12.92}          & \multicolumn{1}{c|}{7.01}          & \multicolumn{1}{c|}{10.63}          &  18.60  \\                                                                                          
\multirow{-3}{*}{Amac}                        & RITR                        & \multicolumn{1}{c|}{\textbf{4.26}}  & \multicolumn{1}{c|}{\textbf{7.47}}  & \multicolumn{1}{c|}{\textbf{15.72}} & \multicolumn{1}{c|}{\textbf{10.44}} & \multicolumn{1}{c|}{\textbf{14.73}} & \textbf{23.50} \\ \hline
                                              & RITR w/o STC                            & \multicolumn{1}{c|}{\underline{3.96}}           & \multicolumn{1}{c|}{\underline{7.16}}           & \multicolumn{1}{c|}{\underline{15.58}}          & \multicolumn{1}{c|}{\underline{10.04}}          & \multicolumn{1}{c|}{\underline{14.35}}          & \underline{23.33}          \\
                                              & RITR w/o ITR                            & \multicolumn{1}{c|}{3.04}           & \multicolumn{1}{c|}{5.79}          & \multicolumn{1}{c|}{13.46}          & \multicolumn{1}{c|}{7.53}          & \multicolumn{1}{c|}{11.25}          &  19.47      \\   
\multirow{-3}{*}{Amap}                        & RITR                        & \multicolumn{1}{c|}{\textbf{4.25}}  & \multicolumn{1}{c|}{\textbf{7.50}}  & \multicolumn{1}{c|}{\textbf{16.00}} & \multicolumn{1}{c|}{\textbf{10.42}} & \multicolumn{1}{c|}{\textbf{14.78}} & \textbf{23.84} \\ \hline
\end{tabular}
\label{VII}
\end{table*}

\subsection{Hybrid-absent Case}
\subsubsection{Performance Comparison (Q1)}
In this section, we further investigate the performance of our proposed methods and study a more challenging hybrid-absent (\textit{i.e.,} marked as ``HA") problem that both attribute-missing and attribute-incomplete samples exist simultaneously within a graph. ``X" or ``X+A" indicates that the classifier receives the attribute matrix or attribute and adjacency matrices as input in the node classification task. To evaluate the quality of the rebuilt attribute matrix, we take six methods (\textit{i.e.,} GCN, GAT, GINN, GCNMF, SVGA, and SAT) as baselines and report the classification accuracy on four datasets. From these results in Table \ref{VI}, we can find that 1) although ITR outperforms all baseline methods and achieves the most competitive results, it suffers from a significant performance degradation by 4.37\% on average in the ``HA"(``X") case compared to that in the ``AM"(``X") case. This is because ITR conducts sample embedding over attribute-incomplete samples directly so that amounts of error information have diffused through the network. As a result, the resultant representations are inaccurate and can hardly provide the attribute-missing samples with discriminative enough information for feature completion; 2) taking the ``HA"(``X") case for example, RITR achieves the best performance against all compared baselines. Specifically, RITR improves SVGA and SAT by 10.20\%/5.58\%, 9.24\%/9.60\%, 18.48\%/13.98\%, and 8.00\%/5.96\% accuracy increment on all datasets. The observations of other cases are similar. These results once again verify that when both attribute-incomplete and attribute-missing samples exist simultaneously, the feature completion mechanisms of these baselines may have an adverse effect on the quality of recovered attributes due to incorrect information diffusion. RITR effectively alleviates this adverse influence by designing two personalized feature completion components with the initializing-then-refining imputation criterion.

\begin{table*}[!t]
\centering
\caption{Performance comparison between SAT and our proposed RITR. We fix the attribute-missing ratio as 60\% and vary the attribute-incomplete ratio (AIR) from 10\% to 70\%. The boldface values indicate the best results (\%).}
\begin{tabular}{c|cccccccccccccc}
\hline
Dateset   & \multicolumn{14}{c}{Cora}  \\ \hline
AIR       & \multicolumn{2}{c|}{{  10\%}}                                                   & \multicolumn{2}{c|}{20\%}                                        & \multicolumn{2}{c|}{30\%}                                        & \multicolumn{2}{c|}{40\%}                                        & \multicolumn{2}{c|}{50\%}                                        & \multicolumn{2}{c|}{60\%}                                        & \multicolumn{2}{c}{70\%}                   \\ \hline
Method & \multicolumn{1}{c|}{{  SAT}} & \multicolumn{1}{c|}{{  RITR}} & \multicolumn{1}{c|}{SAT}   & \multicolumn{1}{c|}{RITR}           & \multicolumn{1}{c|}{SAT}   & \multicolumn{1}{c|}{RITR}           & \multicolumn{1}{c|}{SAT}   & \multicolumn{1}{c|}{RITR}           & \multicolumn{1}{c|}{SAT}   & \multicolumn{1}{c|}{RITR}           & \multicolumn{1}{c|}{SAT}   & \multicolumn{1}{c|}{RITR}           & \multicolumn{1}{c|}{SAT}   & RITR           \\ \hline
Recall@10 & \multicolumn{1}{c|}{14.74}                      & \multicolumn{1}{c|}{\textbf{17.12}}              & \multicolumn{1}{c|}{14.84} & \multicolumn{1}{c|}{\textbf{16.59}} & \multicolumn{1}{c|}{14.23} & \multicolumn{1}{c|}{\textbf{16.08}} & \multicolumn{1}{c|}{13.84} & \multicolumn{1}{c|}{\textbf{15.86}} & \multicolumn{1}{c|}{13.59} & \multicolumn{1}{c|}{\textbf{14.95}} & \multicolumn{1}{c|}{12.62} & \multicolumn{1}{c|}{\textbf{14.36}} & \multicolumn{1}{c|}{12.34} & \textbf{13.16} \\ 
Recall@20 & \multicolumn{1}{c|}{21.45}                      & \multicolumn{1}{c|}{\textbf{23.72}}              & \multicolumn{1}{c|}{21.46} & \multicolumn{1}{c|}{\textbf{23.36}} & \multicolumn{1}{c|}{20.70} & \multicolumn{1}{c|}{\textbf{23.02}} & \multicolumn{1}{c|}{20.17} & \multicolumn{1}{c|}{\textbf{22.29}} & \multicolumn{1}{c|}{19.57} & \multicolumn{1}{c|}{\textbf{21.40}} & \multicolumn{1}{c|}{18.46} & \multicolumn{1}{c|}{\textbf{20.69}} & \multicolumn{1}{c|}{16.78} & \textbf{18.72} \\ 
Recall@50 & \multicolumn{1}{c|}{34.08}                      & \multicolumn{1}{c|}{\textbf{36.32}}              & \multicolumn{1}{c|}{34.00} & \multicolumn{1}{c|}{\textbf{35.76}} & \multicolumn{1}{c|}{33.06} & \multicolumn{1}{c|}{\textbf{35.19}} & \multicolumn{1}{c|}{32.34} & \multicolumn{1}{c|}{\textbf{34.38}} & \multicolumn{1}{c|}{31.48} & \multicolumn{1}{c|}{\textbf{33.38}} & \multicolumn{1}{c|}{29.97} & \multicolumn{1}{c|}{\textbf{32.43}} & \multicolumn{1}{c|}{27.75} & \textbf{30.21} \\ 
NDCG@10   & \multicolumn{1}{c|}{20.70}                      & \multicolumn{1}{c|}{\textbf{23.94}}              & \multicolumn{1}{c|}{20.73} & \multicolumn{1}{c|}{\textbf{23.35}} & \multicolumn{1}{c|}{20.12} & \multicolumn{1}{c|}{\textbf{22.66}} & \multicolumn{1}{c|}{19.50} & \multicolumn{1}{c|}{\textbf{22.37}} & \multicolumn{1}{c|}{19.12} & \multicolumn{1}{c|}{\textbf{21.37}} & \multicolumn{1}{c|}{18.01} & \multicolumn{1}{c|}{\textbf{20.51}} & \multicolumn{1}{c|}{16.83} & \textbf{19.08} \\ 
NDCG@20   & \multicolumn{1}{c|}{25.11}                      & \multicolumn{1}{c|}{\textbf{28.40}}              & \multicolumn{1}{c|}{25.09} & \multicolumn{1}{c|}{\textbf{27.86}} & \multicolumn{1}{c|}{24.41} & \multicolumn{1}{c|}{\textbf{27.26}} & \multicolumn{1}{c|}{23.74} & \multicolumn{1}{c|}{\textbf{26.67}} & \multicolumn{1}{c|}{23.11} & \multicolumn{1}{c|}{\textbf{25.69}} & \multicolumn{1}{c|}{21.91} & \multicolumn{1}{c|}{\textbf{24.69}} & \multicolumn{1}{c|}{19.83} & \textbf{22.80} \\ 
NDCG@50   & \multicolumn{1}{c|}{31.80}                      & \multicolumn{1}{c|}{\textbf{35.10}}              & \multicolumn{1}{c|}{31.78} & \multicolumn{1}{c|}{\textbf{34.44}} & \multicolumn{1}{c|}{30.90} & \multicolumn{1}{c|}{\textbf{33.73}} & \multicolumn{1}{c|}{30.15} & \multicolumn{1}{c|}{\textbf{33.06}} & \multicolumn{1}{c|}{29.40} & \multicolumn{1}{c|}{\textbf{32.04}} & \multicolumn{1}{c|}{27.94} & \multicolumn{1}{c|}{\textbf{30.88}} & \multicolumn{1}{c|}{25.62} & \textbf{28.83} \\ \hline
Dateset   & \multicolumn{14}{c}{Citeseer}  \\ \hline
AIR       & \multicolumn{2}{c|}{10\%}                                                                          & \multicolumn{2}{c|}{20\%}                                        & \multicolumn{2}{c|}{30\%}                                        & \multicolumn{2}{c|}{40\%}                                        & \multicolumn{2}{c|}{50\%}                                        & \multicolumn{2}{c|}{60\%}                                        & \multicolumn{2}{c}{70\%}                   \\ \hline
Method & \multicolumn{1}{c|}{SAT}                        & \multicolumn{1}{c|}{RITR}                        & \multicolumn{1}{c|}{SAT}   & \multicolumn{1}{c|}{RITR}           & \multicolumn{1}{c|}{SAT}   & \multicolumn{1}{c|}{RITR}           & \multicolumn{1}{c|}{SAT}   & \multicolumn{1}{c|}{RITR}           & \multicolumn{1}{c|}{SAT}   & \multicolumn{1}{c|}{RITR}           & \multicolumn{1}{c|}{SAT}   & \multicolumn{1}{c|}{RITR}           & \multicolumn{1}{c|}{SAT}   & RITR           \\ \hline
Recall@10 & \multicolumn{1}{c|}{7.27}                       & \multicolumn{1}{c|}{\textbf{9.69}}               & \multicolumn{1}{c|}{7.23}  & \multicolumn{1}{c|}{\textbf{9.48}}  & \multicolumn{1}{c|}{7.24}  & \multicolumn{1}{c|}{\textbf{9.09}}  & \multicolumn{1}{c|}{6.95}  & \multicolumn{1}{c|}{\textbf{8.79}}  & \multicolumn{1}{c|}{6.61}  & \multicolumn{1}{c|}{\textbf{8.44}}  & \multicolumn{1}{c|}{5.84}  & \multicolumn{1}{c|}{\textbf{7.78}}  & \multicolumn{1}{c|}{5.68}  & \textbf{7.19}  \\ 
Recall@20 & \multicolumn{1}{c|}{12.33}                      & \multicolumn{1}{c|}{\textbf{15.50}}              & \multicolumn{1}{c|}{11.95} & \multicolumn{1}{c|}{\textbf{15.12}} & \multicolumn{1}{c|}{11.94} & \multicolumn{1}{c|}{\textbf{14.59}} & \multicolumn{1}{c|}{11.52} & \multicolumn{1}{c|}{\textbf{14.00}} & \multicolumn{1}{c|}{11.03} & \multicolumn{1}{c|}{\textbf{13.44}} & \multicolumn{1}{c|}{9.95}  & \multicolumn{1}{c|}{\textbf{12.51}} & \multicolumn{1}{c|}{9.97}  & \textbf{11.54} \\ 
Recall@50 & \multicolumn{1}{c|}{22.90}                      & \multicolumn{1}{c|}{\textbf{26.36}}              & \multicolumn{1}{c|}{22.28} & \multicolumn{1}{c|}{\textbf{25.87}} & \multicolumn{1}{c|}{22.55} & \multicolumn{1}{c|}{\textbf{25.16}} & \multicolumn{1}{c|}{21.75} & \multicolumn{1}{c|}{\textbf{24.23}} & \multicolumn{1}{c|}{20.88} & \multicolumn{1}{c|}{\textbf{23.46}} & \multicolumn{1}{c|}{19.20} & \multicolumn{1}{c|}{\textbf{22.14}} & \multicolumn{1}{c|}{18.99} & \textbf{20.58} \\ 
NDCG@10   & \multicolumn{1}{c|}{12.61}                      & \multicolumn{1}{c|}{\textbf{16.50}}              & \multicolumn{1}{c|}{12.40} & \multicolumn{1}{c|}{\textbf{16.16}} & \multicolumn{1}{c|}{12.25} & \multicolumn{1}{c|}{\textbf{15.59}} & \multicolumn{1}{c|}{11.85} & \multicolumn{1}{c|}{\textbf{15.11}} & \multicolumn{1}{c|}{11.29} & \multicolumn{1}{c|}{\textbf{14.47}} & \multicolumn{1}{c|}{9.71}  & \multicolumn{1}{c|}{\textbf{13.38}} & \multicolumn{1}{c|}{8.63}  & \textbf{12.51} \\ 
NDCG@20   & \multicolumn{1}{c|}{16.83}                      & \multicolumn{1}{c|}{\textbf{21.37}}              & \multicolumn{1}{c|}{16.34} & \multicolumn{1}{c|}{\textbf{20.88}} & \multicolumn{1}{c|}{16.17} & \multicolumn{1}{c|}{\textbf{20.18}} & \multicolumn{1}{c|}{15.65} & \multicolumn{1}{c|}{\textbf{19.46}} & \multicolumn{1}{c|}{14.98} & \multicolumn{1}{c|}{\textbf{18.64}} & \multicolumn{1}{c|}{13.13} & \multicolumn{1}{c|}{\textbf{17.33}} & \multicolumn{1}{c|}{12.21} & \textbf{16.14} \\ 
NDCG@50   & \multicolumn{1}{c|}{23.74}                      & \multicolumn{1}{c|}{\textbf{28.50}}              & \multicolumn{1}{c|}{23.09} & \multicolumn{1}{c|}{\textbf{27.94}} & \multicolumn{1}{c|}{23.11} & \multicolumn{1}{c|}{\textbf{27.13}} & \multicolumn{1}{c|}{22.35} & \multicolumn{1}{c|}{\textbf{26.18}} & \multicolumn{1}{c|}{21.41} & \multicolumn{1}{c|}{\textbf{25.21}} & \multicolumn{1}{c|}{19.19} & \multicolumn{1}{c|}{\textbf{23.67}} & \multicolumn{1}{c|}{18.10} & \textbf{22.06} \\ \hline
Dateset   & \multicolumn{14}{c}{Amac}  \\ \hline
AIR       & \multicolumn{2}{c|}{10\%}                                                                          & \multicolumn{2}{c|}{20\%}                                        & \multicolumn{2}{c|}{30\%}                                        & \multicolumn{2}{c|}{40\%}                                        & \multicolumn{2}{c|}{50\%}                                        & \multicolumn{2}{c|}{60\%}                                        & \multicolumn{2}{c}{70\%}                   \\ \hline
Method & \multicolumn{1}{c|}{SAT}                        & \multicolumn{1}{c|}{RITR}                        & \multicolumn{1}{c|}{SAT}   & \multicolumn{1}{c|}{RITR}           & \multicolumn{1}{c|}{SAT}   & \multicolumn{1}{c|}{RITR}           & \multicolumn{1}{c|}{SAT}   & \multicolumn{1}{c|}{RITR}           & \multicolumn{1}{c|}{SAT}   & \multicolumn{1}{c|}{RITR}           & \multicolumn{1}{c|}{SAT}   & \multicolumn{1}{c|}{RITR}           & \multicolumn{1}{c|}{SAT}   & RITR           \\ \hline
Recall@10 & \multicolumn{1}{c|}{3.81}                       & \multicolumn{1}{c|}{\textbf{4.50}}               & \multicolumn{1}{c|}{3.81}  & \multicolumn{1}{c|}{\textbf{4.47}}  & \multicolumn{1}{c|}{3.78}  & \multicolumn{1}{c|}{\textbf{4.44}}  & \multicolumn{1}{c|}{3.75}  & \multicolumn{1}{c|}{\textbf{4.39}}  & \multicolumn{1}{c|}{3.70}  & \multicolumn{1}{c|}{\textbf{4.36}}  & \multicolumn{1}{c|}{3.59}  & \multicolumn{1}{c|}{\textbf{4.26}}  & \multicolumn{1}{c|}{3.24}  & \textbf{4.14}  \\ 
Recall@20 & \multicolumn{1}{c|}{6.88}                       & \multicolumn{1}{c|}{\textbf{7.88}}               & \multicolumn{1}{c|}{6.82}  & \multicolumn{1}{c|}{\textbf{7.83}}  & \multicolumn{1}{c|}{6.79}  & \multicolumn{1}{c|}{\textbf{7.76}}  & \multicolumn{1}{c|}{6.70}  & \multicolumn{1}{c|}{\textbf{7.69}}  & \multicolumn{1}{c|}{6.69}  & \multicolumn{1}{c|}{\textbf{7.60}}  & \multicolumn{1}{c|}{6.52}  & \multicolumn{1}{c|}{\textbf{7.47}}  & \multicolumn{1}{c|}{6.13}  & \textbf{7.39}  \\ 
Recall@50 & \multicolumn{1}{c|}{14.86}                      & \multicolumn{1}{c|}{\textbf{16.24}}              & \multicolumn{1}{c|}{14.79} & \multicolumn{1}{c|}{\textbf{16.21}} & \multicolumn{1}{c|}{14.73} & \multicolumn{1}{c|}{\textbf{16.10}} & \multicolumn{1}{c|}{14.64} & \multicolumn{1}{c|}{\textbf{16.02}} & \multicolumn{1}{c|}{14.52} & \multicolumn{1}{c|}{\textbf{15.91}} & \multicolumn{1}{c|}{14.28} & \multicolumn{1}{c|}{\textbf{15.72}} & \multicolumn{1}{c|}{13.99} & \textbf{15.83} \\ 
NDCG@10   & \multicolumn{1}{c|}{9.38}                       & \multicolumn{1}{c|}{\textbf{10.97}}              & \multicolumn{1}{c|}{9.37}  & \multicolumn{1}{c|}{\textbf{10.94}} & \multicolumn{1}{c|}{9.32}  & \multicolumn{1}{c|}{\textbf{10.87}} & \multicolumn{1}{c|}{9.26}  & \multicolumn{1}{c|}{\textbf{10.75}} & \multicolumn{1}{c|}{9.09}  & \multicolumn{1}{c|}{\textbf{10.65}} & \multicolumn{1}{c|}{8.86}  & \multicolumn{1}{c|}{\textbf{10.44}} & \multicolumn{1}{c|}{8.75}  & \textbf{10.17} \\ 
NDCG@20   & \multicolumn{1}{c|}{13.49}                      & \multicolumn{1}{c|}{\textbf{15.45}}              & \multicolumn{1}{c|}{13.42} & \multicolumn{1}{c|}{\textbf{15.39}} & \multicolumn{1}{c|}{13.36} & \multicolumn{1}{c|}{\textbf{15.29}} & \multicolumn{1}{c|}{13.23} & \multicolumn{1}{c|}{\textbf{15.14}} & \multicolumn{1}{c|}{13.09} & \multicolumn{1}{c|}{\textbf{14.98}} & \multicolumn{1}{c|}{12.79} & \multicolumn{1}{c|}{\textbf{14.73}} & \multicolumn{1}{c|}{12.40} & \textbf{14.53} \\ 
NDCG@50   & \multicolumn{1}{c|}{22.00}                      & \multicolumn{1}{c|}{\textbf{24.33}}              & \multicolumn{1}{c|}{21.92} & \multicolumn{1}{c|}{\textbf{24.30}} & \multicolumn{1}{c|}{21.84} & \multicolumn{1}{c|}{\textbf{24.16}} & \multicolumn{1}{c|}{21.70} & \multicolumn{1}{c|}{\textbf{23.99}} & \multicolumn{1}{c|}{21.47} & \multicolumn{1}{c|}{\textbf{23.81}} & \multicolumn{1}{c|}{21.11} & \multicolumn{1}{c|}{\textbf{23.50}}  & \multicolumn{1}{c|}{20.91} & \textbf{23.53} \\ \hline
Dateset   & \multicolumn{14}{c}{Amap}  \\ \hline
AIR       & \multicolumn{2}{c|}{10\%}                                                                          & \multicolumn{2}{c|}{20\%}                                        & \multicolumn{2}{c|}{30\%}                                        & \multicolumn{2}{c|}{40\%}                                        & \multicolumn{2}{c|}{50\%}                                        & \multicolumn{2}{c|}{60\%}                                        & \multicolumn{2}{c}{70\%}                   \\ \hline
Method & \multicolumn{1}{c|}{SAT}                        & \multicolumn{1}{c|}{RITR}                        & \multicolumn{1}{c|}{SAT}   & \multicolumn{1}{c|}{RITR}           & \multicolumn{1}{c|}{SAT}   & \multicolumn{1}{c|}{RITR}           & \multicolumn{1}{c|}{SAT}   & \multicolumn{1}{c|}{RITR}           & \multicolumn{1}{c|}{SAT}   & \multicolumn{1}{c|}{RITR}           & \multicolumn{1}{c|}{SAT}   & \multicolumn{1}{c|}{RITR}           & \multicolumn{1}{c|}{SAT}   & RITR           \\ \hline
Recall@10 & \multicolumn{1}{c|}{3.95}                       & \multicolumn{1}{c|}{\textbf{4.40}}               & \multicolumn{1}{c|}{3.90}  & \multicolumn{1}{c|}{\textbf{4.37}}  & \multicolumn{1}{c|}{3.92}  & \multicolumn{1}{c|}{\textbf{4.35}}  & \multicolumn{1}{c|}{3.90}  & \multicolumn{1}{c|}{\textbf{4.30}}  & \multicolumn{1}{c|}{3.86}  & \multicolumn{1}{c|}{\textbf{4.27}}  & \multicolumn{1}{c|}{3.82}  & \multicolumn{1}{c|}{\textbf{4.25}}  & \multicolumn{1}{c|}{3.53}  & \textbf{4.12}  \\ 
Recall@20 & \multicolumn{1}{c|}{7.16}                       & \multicolumn{1}{c|}{\textbf{7.86}}               & \multicolumn{1}{c|}{7.13}  & \multicolumn{1}{c|}{\textbf{7.82}}  & \multicolumn{1}{c|}{7.13}  & \multicolumn{1}{c|}{\textbf{7.76}}  & \multicolumn{1}{c|}{7.05}  & \multicolumn{1}{c|}{\textbf{7.71}}  & \multicolumn{1}{c|}{6.97}  & \multicolumn{1}{c|}{\textbf{7.64}}  & \multicolumn{1}{c|}{6.93}  & \multicolumn{1}{c|}{\textbf{7.50}}  & \multicolumn{1}{c|}{6.43}  & \textbf{7.26}  \\ 
Recall@50 & \multicolumn{1}{c|}{15.49}                      & \multicolumn{1}{c|}{\textbf{16.48}}              & \multicolumn{1}{c|}{15.42} & \multicolumn{1}{c|}{\textbf{16.44}} & \multicolumn{1}{c|}{15.44} & \multicolumn{1}{c|}{\textbf{16.36}} & \multicolumn{1}{c|}{15.34} & \multicolumn{1}{c|}{\textbf{16.29}} & \multicolumn{1}{c|}{15.16} & \multicolumn{1}{c|}{\textbf{16.19}} & \multicolumn{1}{c|}{15.20} & \multicolumn{1}{c|}{\textbf{16.00}} & \multicolumn{1}{c|}{14.14} & \textbf{15.39} \\ 
NDCG@10   & \multicolumn{1}{c|}{9.68}                       & \multicolumn{1}{c|}{\textbf{10.80}}               & \multicolumn{1}{c|}{9.60}  & \multicolumn{1}{c|}{\textbf{10.75}} & \multicolumn{1}{c|}{9.63}  & \multicolumn{1}{c|}{\textbf{10.67}} & \multicolumn{1}{c|}{9.57}  & \multicolumn{1}{c|}{\textbf{10.57}} & \multicolumn{1}{c|}{9.48}  & \multicolumn{1}{c|}{\textbf{10.48}} & \multicolumn{1}{c|}{9.39}  & \multicolumn{1}{c|}{\textbf{10.42}} & \multicolumn{1}{c|}{8.71}  & \textbf{10.14} \\ 
NDCG@20   & \multicolumn{1}{c|}{13.98}                      & \multicolumn{1}{c|}{\textbf{15.39}}              & \multicolumn{1}{c|}{13.92} & \multicolumn{1}{c|}{\textbf{15.33}} & \multicolumn{1}{c|}{13.93} & \multicolumn{1}{c|}{\textbf{15.22}} & \multicolumn{1}{c|}{13.81} & \multicolumn{1}{c|}{\textbf{15.11}} & \multicolumn{1}{c|}{13.65} & \multicolumn{1}{c|}{\textbf{14.96}} & \multicolumn{1}{c|}{13.57} & \multicolumn{1}{c|}{\textbf{14.78}} & \multicolumn{1}{c|}{12.62} & \textbf{14.34} \\ 
NDCG@50   & \multicolumn{1}{c|}{22.86}                      & \multicolumn{1}{c|}{\textbf{24.55}}              & \multicolumn{1}{c|}{22.76} & \multicolumn{1}{c|}{\textbf{24.50}} & \multicolumn{1}{c|}{22.80} & \multicolumn{1}{c|}{\textbf{24.37}} & \multicolumn{1}{c|}{22.66} & \multicolumn{1}{c|}{\textbf{24.25}} & \multicolumn{1}{c|}{22.41} & \multicolumn{1}{c|}{\textbf{24.09}} & \multicolumn{1}{c|}{22.39} & \multicolumn{1}{c|}{\textbf{23.84}} & \multicolumn{1}{c|}{20.89} & \textbf{23.00} \\ \hline
\end{tabular}
\label{VIII}
\end{table*}

\begin{figure*}[!t]
\centering
\includegraphics[width=7.1in]{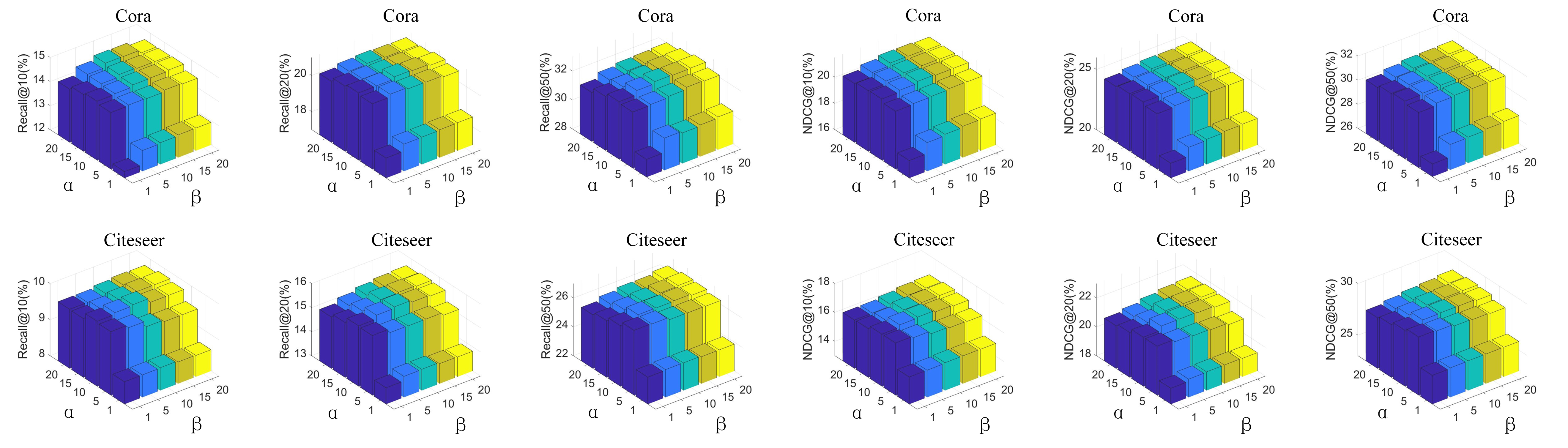}
\caption{The sensitivity analysis of RITR with the variation of two hyper-parameters. Both attribute-incomplete and attribute-missing ratios are set to 60\%.}
\label{5}
\end{figure*}

\begin{figure*}[!t]
\centering
\includegraphics[width=6.8in]{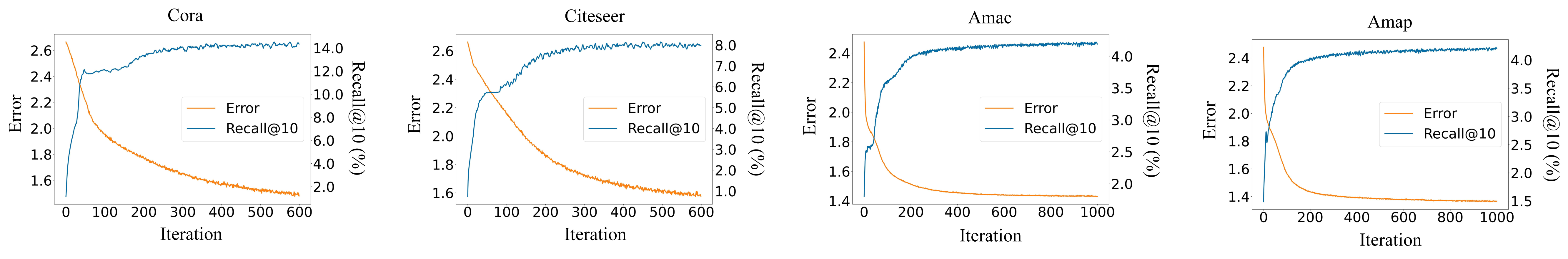}
\caption{Illustration of method convergence and performance variation of RITR. X-axis, left Y-axis, and right Y-axis refer to the iteration number, the final objective error, and the Recall@10 performance, respectively. Both attribute-incomplete and attribute-missing ratios are set to 60\%.}
\label{6}
\end{figure*}

\subsubsection{Effect of Each Component of RITR (Q2)}
Here we conduct an ablation study to validate the effectiveness of our proposed attribute-incomplete and attribute-missing imputation mechanisms. Table \ref{VII} reports the Recall and NDCG performance of three methods, including RITR w/o STC, RITR w/o ITR, and RITR. Specially, RITR w/o STC or RITR w/o ITR indicates that the method removes the sample-denoising then consistency-preserving component or the initializing then refining component. Note that here we set both attribute-missing and incomplete ratios as 60\%. Table \ref{VII} compares the results of RITR and its two variants, from which we can see that 1) RITR consistently improves RITR w/o STC on all datasets. Taking the results on Cora for instance, RITR gains 0.73\%, 1.02\%, 1.52\%, 1.06\%, 1.22\%, and 1.45\% increment in terms of Recall and NDCG, demonstrating the effectiveness of leveraging the intimate structure-attribute relationship to guide the imputation of incomplete attributes. Similar observations can be concluded from the results on other datasets; 2) RITR significantly outperforms RITR w/o ITR and has performance enhancements of 4.73\%, 3.97\%, 2.80\%, and 2.54\% over it in terms of Recall@50 on four datasets, respectively. These results imply the importance of an effective imputation strategy in which we employ the most trustworthy visible information to implement the missing attribute completion. In summary, this ablation study clearly validates that each component can contribute to the overall performance of RITR.

\subsubsection{Analysis of The Attribute-incomplete Ratio (Q3)}
To further investigate the superiority of RITR, it is necessary to show whether the proposed RITR can still achieve effective feature completion when less visible attribute information is available. To this end, we make a performance comparison between SAT and RITR by varying the attribute-incomplete ratio from 10\% to 70\% and fixing the attribute-missing ratio as 60\%. From the results in Table \ref{VIII}, several observations can be summarized: 1) RITR consistently performs better than SAT in all situations on four datasets. For example, RITR outperforms SAT by 2.42\%, 2.25\%, 1.85\%, 1.84\%, 1.83\%, 1.94\%, and 1.51\% in terms of Recall@10 when the attribute-incomplete ratio varies from 10\% to 70\% on Citeseer. The observations of other metrics and datasets are similar. This is because SAT can not implement an effective latent distribution matching between incomplete attributes and structures. Naturally, the resultant misleading information poses a negative impact on data imputation and feature completion, resulting in sub-optimal node representations. RITR can effectively model hybrid-absent graphs and alleviate the diffusion of inaccurate information under the guidance of initializing-then-refining imputation criterion; 2) taking the results of Recall@10/NDCG@10 on Amac and Amap for example, RITR with 70\% incomplete attributes can still achieve better performance than SAT with 10\% ones. These results illustrate that RITR can still achieve high-quality data imputation and feature completion with limited observed signals. Overall, all the above results solidly demonstrate the superiority and robustness of RITR.  

\subsubsection{Hyper-parameter Analysis (Q4)}
As seen in Eq.(\ref{eq:13}), RITR introduces two hyper-parameters to balance the importance of different objectives. To show their influence in depth, we conduct an experiment to investigate the effect of $\alpha$ and $\beta$. Note that we first set one to a certain value and then tune the other carefully. Fig. \ref{5} reports the Recall and NDCG performance variation of RITR on Cora and Citeseer when $\alpha$ and $\beta$ vary from 1 to 20 with a step size of 5. From these sub-figures, we can observe that 1) tuning both $\alpha$ and $\beta$ would cause performance variation and the model performance is more stable in the range of [5, 15], suggesting that searching $\alpha$ and $\beta$ values from a reasonable hyper-parameter region could benefit the model performance; 2) for a certain $\alpha$ value, the performance shows a trend of first rising and then dropping slightly with the variation of $\beta$. This indicates that RITR needs a proper coefficient to guarantee the structure-attribute consistency for improving the quality of feature completion. As shown, the performance of the model with a certain $\beta$ value has similar trends when we change the $\alpha$ value; 3) RITR tends to perform well by setting $\alpha$ and $\beta$ to 10 according to the results of all datasets.

\subsubsection{Convergence and Performance Variation (Q5)}
To illustrate the convergence of the proposed RITR, we record the profiling performance reflected by the Recall@10 metric and plot the objective error of RITR with iterations on four datasets. From these sub-figures illustrated in Fig. \ref{6}, we can observe that 1) the Recall@10 metric of RITR first gradually increases to a plateau with an obvious tendency and then keeps stable with a wide range of iterations; 2) RITR can converge within 1000 epochs on four datasets. These results clearly verify the good convergence property of our proposed method and reveal the effectiveness of the learning procedure.

\section{Conclusion and Future Work}
Hybrid-absent graphs are ubiquitous in practical applications. However, the corresponding learning problem that significantly influences the performance of existing graph machine learning methods is still left under-explored. We firstly propose ITR toward the attribute-missing circumstance, which can enable the attribute and structure information to sufficiently negotiate with each other for accurate missing value restoration. We further improve ITR and design a variant called RITR to handle hybrid-absent graphs, which can effectively leverage the intimate structure-attribute relationship to guide the imputation of incomplete attributes and employ the most trustworthy visible information to implement the missing attribute completion.
Extensive experiments on four benchmark datasets have been conducted to compare two proposed methods with state-of-the-art competitors. These results have solidly demonstrated the superiority and robustness of ITR and RITR on both profiling and node classification tasks. However, some limitations of existing attribute-missing or hybrid-absent graph machine learning methods are we still have not explored some issues for them. For instance, the time complexities of most methods are $\mathcal{O}(N^2)$, making them hard to be deployed to various large-scale graph-oriented applications. Future work may extend the proposed RITR to a scalable version with linear scalability (\textit{i.e., $\mathcal{O}(BNd)$}) via a mini-batch design. 
Moreover, in the current version, RITR conducts the structure-attribute information interaction and imputation via a simple concatenation. In the future, how to develop a more mathematical hybrid-absent graph machine learning approach to illustrate the structure-attribute relationship in theory and alleviate the mutual interference between them is another interesting direction, which may further improve the quality of data imputation and graph representations.

\ifCLASSOPTIONcaptionsoff
  \newpage
\fi

\bibliographystyle{IEEEtran}
\bibliography{mytnnls_ritr}

\end{document}